\documentclass[11pt, copyright, logo, secondlogo]{template/lark}
\renewcommand{\today}{}  
\usepackage[authoryear, sort&compress, round]{natbib}
\usepackage{enumitem}
\setlist{nosep}
\usepackage{xspace}
\usepackage{graphicx}
\usepackage{tabularx}
\usepackage{booktabs}
\usepackage{subcaption}
\usepackage[dvipsnames]{xcolor}
\usepackage{colortbl}
\definecolor{table-blue}{RGB}{173, 216, 230}
\usepackage{hyperref}
\usepackage{verbatim}
\usepackage{float}
\usepackage{cleveref}
\usepackage{wrapfig}
\usepackage{multirow}
\usepackage{fancyvrb}
\usepackage[normalem]{ulem}
\usepackage{url}
\usepackage{tcolorbox}
\usepackage{listings}
\tcbuselibrary{listings,skins,breakable}

\usepackage{amsmath,amsfonts,bm}

\def\eqref#1{equation~\ref{#1}}

\def\1{\bm{1}}

\def\vc{{\bm{c}}}

\def\vo{{\bm{o}}}

\def\vq{{\bm{q}}}
\def\vr{{\bm{r}}}

\def\vx{{\bm{x}}}

\DeclareMathAlphabet{\mathsfit}{\encodingdefault}{\sfdefault}{m}{sl}
\SetMathAlphabet{\mathsfit}{bold}{\encodingdefault}{\sfdefault}{bx}{n}

\newcommand{\our}{\textsc{CodeScaler}\xspace}

\title{\our: Scaling Code LLM Training and Test-Time Inference via Reward Models}

\author[1]{Xiao Zhu$^{*}$}
\author[1]{Xinyu Zhou$^{*}$}
\author[2,4]{Boyu Zhu}
\author[5]{Hanxu Hu}
\author[6]{Mingzhe Du}
\author[2]{Haotian Zhang}
\author[2]{Huiming Wang$^{\dag}$}
\author[1,3]{Zhijiang Guo$^{\dag}$}

\affil[1]{LARK, HKUST(GZ)}
\affil[2]{Kuaishou Technology}
\affil[3]{HKUST}
\affil[4]{UCL}
\affil[5]{UZH}
\affil[6]{NUS}

\correspondingauthor{Zhijiang Guo (zhijiangguo@hkust-gz.edu.cn), Huiming Wang (huiming\_wang@mymail.sutd.edu.sg)}
\githubpage{https://lark-ai-lab.github.io/codescaler.github.io/}
\modelweight{https://huggingface.co/collections/LARK-Lab/codescaler}

\begin{abstract}

Reinforcement Learning from Verifiable Rewards (RLVR) has driven recent progress in code large language models by leveraging execution-based feedback from unit tests, but its scalability is fundamentally constrained by the availability and reliability of high-quality test cases. We propose \our, a reward model designed to scale both reinforcement learning training and test-time inference for code generation. \our is trained on carefully curated preference data derived from verified code problems and incorporates syntax-aware code extraction and validity-preserving reward shaping to ensure stable and robust optimization. Across four coding benchmarks, \our consistently outperforms execution-based RL by \textbf{+1.55} points on Qwen3-8B-Base and \textbf{+4.23} points on Qwen3-14B-Base. By further scaling to 44K problems with additional synthetic data, \our yields \textbf{+14.64} points improvement over the base model without requiring any test cases. At inference time, \our serves as an effective test-time scaling method, achieving performance comparable to unit test approaches while providing a \textbf{10×} reduction in latency. Moreover, \our surpasses existing reward models on RM-Bench not only in the code domain (\textbf{+3.3} points), but also in general and reasoning domains (\textbf{+2.7} points on average).

\end{abstract}
\begin{document}
\maketitle

\section{Introduction}

As the capability boundaries of Large Language Models (LLMs) continue to expand, their applications in software engineering \citep{hui2024qwen2,qwen3technicalreport,DBLP:journals/corr/abs-2312-13010, DBLP:conf/nips/YangJWLYNP24} and algorithmic problem solving \citep{DBLP:journals/corr/abs-2504-01943, deepcoder2025, DBLP:journals/corr/abs-2501-12948} have grown rapidly. As a highly structured task with executable feedback, code generation has become a central benchmark for evaluating the reasoning ability of LLMs \citep{DBLP:conf/iclr/JainHGLYZWSSS25, DBLP:conf/iclr/WhiteDRPF0SJSDS25}.


The success of reinforcement learning in post-training has accelerated the adoption of Reinforcement Learning from Verifiable Rewards~(RLVR) for training code LLMs~\citep{DBLP:journals/corr/abs-2501-12948, DBLP:conf/icml/GehringZCMCS25, DBLP:journals/corr/abs-2510-18471}. By executing model-generated programs on unit tests within a sandbox environment, RLVR provides binary feedback signals that effectively improve functional correctness. However, RLVR critically depends on high-quality problems with carefully curated test cases, typically sourced from programming contests~\citep{deepcoder2025}. While such data strictly covers problem constraints, it is costly to obtain and limited in scale. Recent work~\citep{DBLP:conf/acl/XuLYZP25, DBLP:journals/corr/abs-2505-21297} has explored synthesizing problems and test cases to address this limitation, but these approaches are often computationally expensive and difficult to validate. 

An appealing alternative is to replace executable rewards with a learned reward model, which assigns dense scalar scores to generated code without requiring execution \citep{DBLP:journals/corr/abs-2507-01352, DBLP:conf/acl/ZengJ0NCC25, DBLP:conf/acl/LiDL0WT025}. However, prior attempts to apply reward models to code reinforcement learning have struggled to match the performance of execution-based RLVR, often suffering from training instability and reward hacking \citep{DBLP:conf/acl/ZengJ0NCC25}. As a result, the potential of code reward models remains largely unexplored.


In this work, we propose \our, a reward model designed to scale both code LLM training and test-time inference. \our is trained on high-quality preference data constructed from on-policy RL trajectories, enabling it to reliably distinguish code correctness. We further introduce syntax-aware code extraction and validity-preserving reward shaping to stabilize RM-based RL training. Across four coding benchmarks, \our consistently outperforms execution-based RL by \textbf{+1.55} points on Qwen3-8B-Base and \textbf{+4.23} points on Qwen3-14B-Base. By further scaling to 44K problems with additional synthetic data, \our yields \textbf{+14.64} points improvement over the base model without requiring any test cases. At test time, \our achieves performance comparable to unit test TTS methods while providing a \textbf{10$\times$} speedup, and surpasses prior reward models on RM-Bench across code (\textbf{+3.3}), reasoning, and general domains (\textbf{+2.7} on average). Our contributions are:

$\bullet$ We construct high-quality preference data from on-policy RL trajectories and train \our to reliably distinguish code correctness. We further introduce syntax-aware code extraction and validity-preserving reward shaping to stabilize RM-based RL training.

$\bullet$ We demonstrate that \our serves as an effective reward for both RL training and test-time Best-of-N selection, consistently outperforming RLVR and existing reward models across diverse datasets and model scales.

$\bullet$ We conduct comprehensive ablation studies and analyses that validate each component of \our and provide insights into reward model design for code.

\section{Motivation}

\begin{figure*}[!t]
    \centering
    \includegraphics[width=\linewidth]{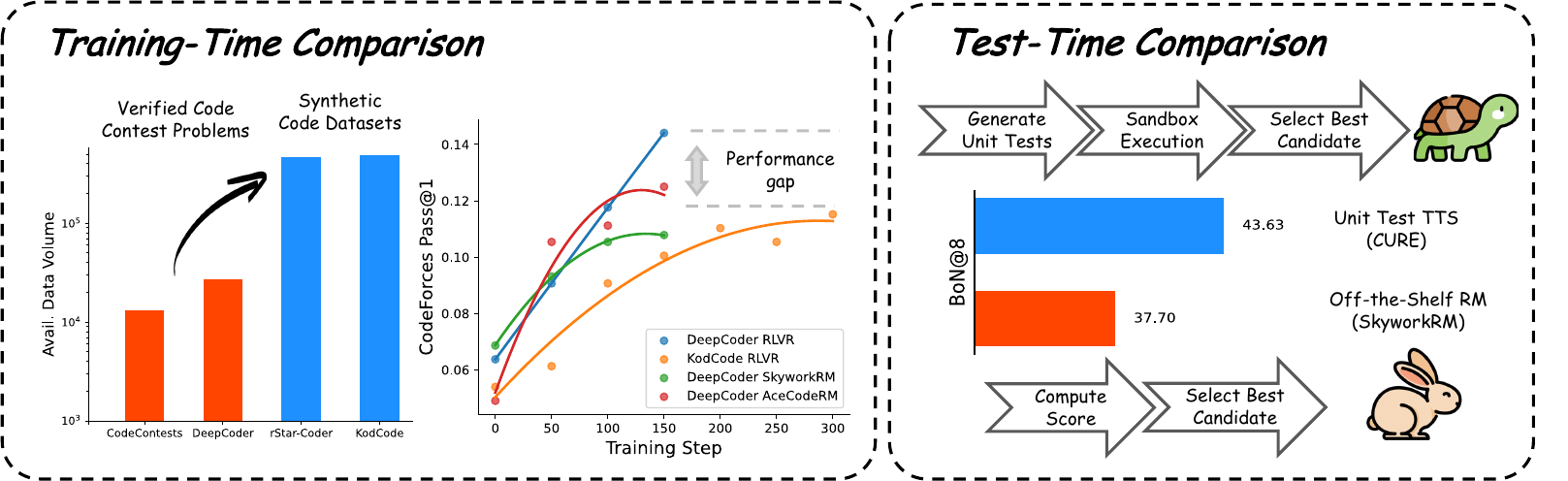}
    \caption{\textit{Left:} Training-Time Comparison shows that despite their larger data scale, synthetic code datasets exhibit a clear performance gap compared to verified code contest problems in RL training. While reward models provide dense supervision, they do not integrate effectively with RL, resulting in weaker performance than RLVR. \textit{Right:} Test-Time Comparison illustrates that Unit Test TTS methods and off-the-shelf reward models demonstrate a clear performance–latency trade-off. This motivates us to develop a reward model that is both effective and efficient for RL training and test-time scaling.}
    \label{fig:motivation}
\end{figure*}

\noindent\textbf{Training-Time Scaling.} RLVR leverages high-quality contest problems with curated test suites and achieves strong performance gains. However, such verified datasets are extremely limited. To address this, recent works construct large-scale synthetic datasets (e.g., rStarCoder \citep{DBLP:journals/corr/abs-2505-21297} with 418K problems and KodCode \citep{DBLP:conf/acl/XuLYZP25} with 447K). However, synthetic problems lack oracle solutions, making it difficult to construct complete and challenging test cases, especially for corner cases. See \autoref{case-study} for a case study.

We trained \textsc{Qwen3-8B-Base} with GRPO on the DeepCoder Dataset (24K high-quality verified problems) \citep{deepcoder2025} and KodCode, respectively. Despite using nearly 2× more data for training, model trained on KodCode dataset fail to outperform that trained on DeepCoder dataset. The DeepCoder-trained model achieves an average performance of 14.41, while the KodCode-trained model reaches only 11.51, resulting in a 2.9 point gap on average (see \autoref{fig:motivation}, left). 

A natural alternative is to use a pretrained RM to provide dense feedback. However, existing RMs are ineffective in practice. As shown in \autoref{fig:motivation} (left), training with \textsc{Skywork-Reward-V2-Qwen3-8B}\footnote{\href{https://huggingface.co/Skywork/Skywork-Reward-V2-Qwen3-8B}{https://huggingface.co/Skywork/Skywork-Reward-V2-Qwen3-8B}. We denote this as \textsc{SkyworkRM} throughout the rest of the paper.} \citep{DBLP:journals/corr/abs-2507-01352} or \textsc{AceCodeRM-7B}\footnote{\href{https://huggingface.co/TIGER-Lab/AceCodeRM-7B}{https://huggingface.co/TIGER-Lab/AceCodeRM-7B}. We denote this as \textsc{AceCodeRM} throughout the rest of the paper.} \citep{DBLP:conf/acl/ZengJ0NCC25} still underperforms standard RLVR even on high-quality DeepCoder data. In detail, RLVR achieves 14.41 average score, whereas \textsc{SkyworkRM} and \textsc{AceCodeRM} achieve only 10.78 and 12.5, respectively. This suggests that current reward models fail to fully exploit high-quality verified training data when directly applied to RL training.

\begin{figure*}[!t]
    \centering
    \includegraphics[width=\linewidth]{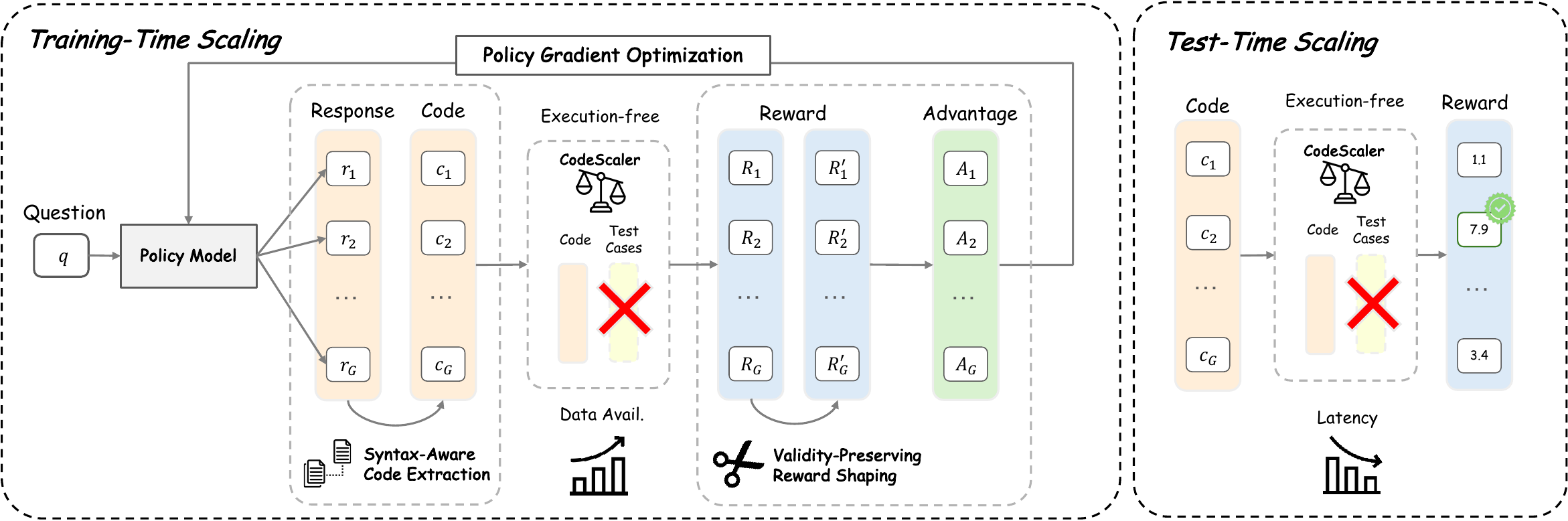}
    \caption{Overall Pipeline of \our for training-time and test-time scaling, which provides execution-free rewards for policy optimization during RL training, and serves as a lightweight sampler for Best-of-N selection, without relying on test-case execution.}
    \label{fig:pipeline}
\end{figure*}

\noindent\textbf{Test-Time Scaling.} Code generation performance can also be improved through test-time scaling techniques such as BoN sampling. A common approach is to generate multiple candidate solutions, execute them against synthesized unit tests, and select the candidate with the highest pass rate \citep{DBLP:journals/corr/abs-2506-03136, li2025s, chencodet}. While effective, such methods require reliable test generation and sandbox execution, resulting in high latency and computational cost. RM can also perform BoN by assigning scalar scores to candidate solutions, using only a single forward pass per candidate, without requiring code execution. 

However, the same limitation of existing RMs also appears at test time. We evaluate \texttt{BoN@8} for a fixed policy model on LiveCodeBench. As shown in \autoref{fig:motivation} (right), the unit test TTS method \textsc{CURE} achieves 43.63 \texttt{BoN@8} on LiveCodeBench, while \textsc{SkyworkRM} achieves only 37.7, resulting in a 5.93 point gap despite its computational advantages, which indicates that current RMs lack sufficient discriminative power to reliably rank candidate solutions.

Taken together, binary execution feedback does not scale with data quantity, while existing RMs fail to serve as effective substitutes in either reinforcement learning or test-time scaling. These challenges motivate the development of a code-specific reward model that can (1) enable execution-free reinforcement learning at scale, and (2) support efficient and effective test-time scaling without relying on test cases.

\section{Method}
In this section, we introduce \our, our proposed reward model designed for Code LLMs. We first describe how we train \our using high-quality code problems, and then explain how \our is effectively applied to downstream RL training and test-time scaling.

\subsection{Reward Model Training}
\subsubsection{Preliminary}

\noindent\textbf{Training Objective.} We aim to train a reward model $r_{\phi}(\cdot)$ that assigns scores to LLM generated code. The model is trained on pairwise preferences with Bradley-Terry loss \citep{bradley1952rank}. Let $\vq$ denote a code problem, and let $\vc^{+}$ and $\vc^{-}$ be two candidates code solutions for the same problem, where $\vc^{+}$ is preferred over $\vc^{-}$. The probability that $\vc^{+}$ is preferred over $\vc^{-}$ is modeled using the Bradley-Terry loss:
\begin{equation}
P(\vc^{+} \succ \vc^{-}|\vq) = \sigma(r_{\phi}(\vq, \vc^{+}) - r_{\phi}(\vq, \vc^{-}))
\end{equation}
where $\sigma(\cdot)$ denotes the sigmoid function. Then the reward model is trained by minimizing the negative log-likelihood over the preference dataset $\mathcal{D}$:
\begin{equation}
\mathcal{L}_{\text{RM}} = -\mathbb{E}_{(\vq, \vc^{+}, \vc^{-})\sim\mathcal{D}}[\log{\sigma(r_{\phi}(\vq, \vc^{+}) - r_{\phi}(\vq, \vc^{-}))}]
\end{equation}


\subsubsection{Preference Dataset Curation}

Prior work highlights the importance of high-quality preference data, showing that a relatively small but carefully curated dataset can suffice to train strong reward models \citep{DBLP:journals/corr/abs-2502-03387}. As noted in DeepCoder \citep{deepcoder2025}, datasets that are overly easy or contain noisy or unverifiable problems can significantly degrade performance. In particular, flawed or low-coverage test cases fail to reliably distinguish solution quality, resulting in noisy preference pairs. To mitigate this issue, we adopt the DeepCoder dataset as seed data for preference construction, which contains 24K high-quality problems with verified test cases. 



\noindent\textbf{Utilizing On-Policy Training Rollouts.} Instead of simply prompting multiple open-source LLMs to generate candidate solutions, we collect training data from a policy model trained with RLVR. Specifically, we train \textsc{Qwen3-8B-Base} on the DeepCoder dataset using GRPO and reuse the resulting on-policy rollouts for reward model construction. As the policy undergoes RL training, its generations naturally span a wide spectrum of solution quality with diversity. 

\noindent\textbf{Preference Pair Construction.} We treat solutions that pass all test cases as positive samples, while solutions that fail any test case are treated as negative samples. \citet{DBLP:conf/acl/ZengJ0NCC25} adopts a threshold-based strategy that labels solutions with a pass rate above 0.8 as correct and enforces a minimum gap between positive and negative samples. In our experiments, we observe that such construction leads to inferior performance in both Best-of-N and RL training (see \autoref{analysis-binary}). To improve the robustness of the RM, we augment the preference pairs by adding misaligned pairs \((\vq, \tilde{\vc})\) as negative, where \(\tilde{\vc}\) is a solution code generated for a different problem. More training details see \autoref{appendix:rm-training-details}.






\subsection{RL Training  with \our}


Most existing code RL pipelines rely on RLVR, where reward is provided as binary feedback through direct code execution \citep{DBLP:conf/icml/GehringZCMCS25, deepcoder2025, DBLP:journals/corr/abs-2510-18471}. In contrast, \our produces more fine-grained, dense reward signals. However, replacing executable rewards with learned rewards also introduces potential challenges, including training instability and susceptibility to reward hacking, as the reward signal is no longer directly grounded in test-case verification. To mitigate these risks, we adopt several techniques when applying \our to RL training.

\noindent\textbf{Syntax-Aware Code Extraction.} In RLVR, code extraction is straightforward as the execution environment naturally filters flawed or syntactically incorrect code. However, RM-based training lacks this inherent verification, making it sensitive to the quality of the input. RM may risk in assigning unpredictably high scores to seemingly plausible but syntactically broken code. To improve the reliability of RM evaluations, we implement a strict code extraction pipeline to extract codes $\vc$ from model responses $\vr$:

1. We extract code only from a single, well-defined code block. Responses containing multiple fragmented code blocks are discarded. Such cases occur primarily during the early stages of RL training, when the policy occasionally attempts to produce partial or incomplete code segments.

2. We perform a static Abstract Syntax Tree (AST) check on extracted code. This ensures that RM only evaluates code that is at least syntactically correct.

Codes $\vc$ that fail any of the above criteria are replaced with the empty string $\epsilon$.

\noindent\textbf{Validity-Preserving Reward Shaping.} The raw output of a Bradley-Terry reward model, $r_{\phi}(\cdot)\in \mathbb{R}$, does not guarantee a global minimum for invalid inputs. To enforce a clear penalty for failed extractions, we define the final reward $R^\prime(\vq, \vc)$ for a given question $\vq$ and extracted code $\vc$ as :
$$R^\prime(\vq, \vc) =
\begin{cases}
f\left(r_{\phi}(\vq, \vc)\right) & \text{if } \vc \text{ is valid} \\
0 & \text{otherwise} 
\end{cases}$$
where $f(z) = \ln(1+e^z) $ is a transformation function that maps RM outputs to a strictly positive range, this ensures that $R^\prime(\vq, \epsilon) < R^\prime(\vq, \vc_{valid})$, thereby providing a consistent optimization target that discourages the generation of unparseable or flawed code.

\subsection{Test-Time Scaling with \our}
We can perform Best-of-N sampling with \our. Given a code problem $\vq$ and a list of candidate code solutions $\vc_1,\vc_2,\cdots,\vc_n$, we assign scores $r_\phi(\vc_1),r_\phi(\vc_2),\cdots,r_\phi(\vc_n)$ using a trained reward model and select the solution with the highest score as the final decision.



\section{Experiments}
We examine the effectiveness of \our in two complementary settings. We first show that \our provides high-quality supervision during RL training, enabling stable and scalable optimization beyond verifiable rewards. We then evaluate \our at inference time, where it serves as an effective ranker for Best-of-N sampling and enables effective and efficient test-time scaling without relying on test-case execution.

\begin{table*}[!t]\small
\centering
\caption{Evaluation results (\texttt{Avg@8}) across four code benchmarks, comparing \our with RLVR on DeepCoder, KodCode, and rStarCoder datasets. \textbf{Bold} indicates the best results. {$\uparrow(\downarrow)$} indicates the improvement (degradation) compared to the RLVR baseline.}
\resizebox{\linewidth}{!}{%
\begin{tabular}{@{}llllll@{}}
\toprule
\multicolumn{1}{l}{\multirow{2}{*}{\textbf{\textsc{Models}}}} & \multicolumn{4}{c}{\textbf{\textsc{Benchmarks}}\texttt{(Avg@8)}}                                                                                                       & \multirow{2}{*}{\textbf{\textsc{Avg.}}} \\
\multicolumn{1}{c}{}                                & \multicolumn{1}{l}{LiveCodeBench} & \multicolumn{1}{l}{CodeContests} & \multicolumn{1}{l}{MBPP} & \multicolumn{1}{l}{CodeForces} &                               \\ \midrule
\multicolumn{6}{c}{\textsc{DeepCoder}}                                                                                                                                                                                    \\ \midrule
\textsc{Qwen3-8B-Base}                                       & 13.75                             & 19.03                            & 61.70                     & 5.35                           & 24.96                          \\
\quad\textit{w/ RLVR}                                    & 23.60                              & 32.00                              & \textbf{76.01}                  & 16.43                        & 37.01                            \\
\quad\textit{w/ \our (Ours)}                                  & \textbf{24.80}\color{red}\textsubscript{(1.20$\uparrow$)}                               & \textbf{33.94}\color{red}\textsubscript{(1.94$\uparrow$)}                             & 75.90\color{ForestGreen}\textsubscript{(0.11$\downarrow$)}                    & \textbf{19.61}\color{red}\textsubscript{(3.27$\uparrow$)}                          & \textbf{38.56}\color{red}\textsubscript{(1.55$\uparrow$)}                          \\ \midrule
\textsc{Qwen3-14B-Base }                                     & 21.37                             & 26.25                            & 71.09                    & 8.45                           & 31.79                            \\
\quad\textit{w/ RLVR}                                    & 25.94                            & 34.30                             & 75.45                   & 20.79                          & 39.12                        \\
\quad\textit{w/ \our (Ours)}                                  & \textbf{27.55}\color{red}\textsubscript{(1.61$\uparrow$)}                             & \textbf{39.33}\color{red}\textsubscript{(5.03$\uparrow$)}                            & \textbf{81.61}\color{red}\textsubscript{(6.16$\uparrow$)}                    & \textbf{24.89}\color{red}\textsubscript{(4.10$\uparrow$)}                         & \textbf{43.35}\color{red}\textsubscript{(4.23$\uparrow$)}                           \\ \midrule
\multicolumn{6}{c}{\textsc{KodCode}}                                                                                                                                                                                      \\ \midrule
\textsc{Qwen3-8B-Base }                                      & 13.75                             & 19.03                            & 61.70                     & 5.35                           & 24.96                          \\
\quad\textit{w/ RLVR}                                    & 21.40                              & 27.87                            & 73.24                   & 11.64                          & 33.54                         \\
\quad\textit{w/ \our (Ours)}                                  & \textbf{23.11}\color{red}\textsubscript{(1.71$\uparrow$)}                             & \textbf{29.34}\color{red}\textsubscript{(1.47$\uparrow$)}                            & \textbf{76.41}\color{red}\textsubscript{(3.17$\uparrow$)}                    & \textbf{12.17}\color{red}\textsubscript{(0.53$\uparrow$)}                          & \textbf{35.26}\color{red}\textsubscript{(1.72$\uparrow$)}                         \\ \midrule
\textsc{Qwen3-14B-Base}                                      & 21.37                             & 26.25                            & 71.09                    & 8.45                           & 31.79                            \\
\quad\textit{w/ RLVR}                                    & 26.12                             & 32.32                            & 78.39                    & 14.10                           & 37.73                         \\
\quad\textit{w/ \our (Ours)}                                  & \textbf{27.78}\color{red}\textsubscript{(1.66$\uparrow$)}                             & \textbf{32.89}\color{red}\textsubscript{(0.57$\uparrow$)}                            & \textbf{81.90}\color{red}\textsubscript{(3.51$\uparrow$)}                    & \textbf{15.17}\color{red}\textsubscript{(1.07$\uparrow$)}                          & \textbf{39.44}\color{red}\textsubscript{(1.71$\uparrow$)}                         \\ \midrule
\multicolumn{6}{c}{\textsc{rStarCoder}}                                                                                                                                                                                   \\ \midrule
\textsc{Qwen3-8B-Base}                                       & 13.75                             & 19.03                            & 61.70                     & 5.35                           & 24.96                          \\
\quad\textit{w/ RLVR}                                    & 23.30                           & 29.44                           & \textbf{75.79}                  & 13.19                         & 35.43                      \\
\quad\textit{w/ \our (Ours)}                                  & \textbf{24.50}\color{red}\textsubscript{(1.20$\uparrow$)}                              & \textbf{30.75}\color{red}\textsubscript{(1.31$\uparrow$)}                            & 73.07\color{ForestGreen}\textsubscript{(2.72$\downarrow$)}                    & \textbf{15.95}\color{red}\textsubscript{(2.76$\uparrow$)}                          & \textbf{36.07}\color{red}\textsubscript{(0.64$\uparrow$)}                         \\ \midrule
\textsc{Qwen3-14B-Base }                                     & 21.37                             & 26.25                            & 71.09                    & 8.45                           & 31.79                            \\
\quad\textit{w/ RLVR}                                    & 27.24                            & \textbf{33.26}                           & 77.54                 & 15.65                       & 38.42                    \\
\quad\textit{w/ \our (Ours) }                                  & \textbf{27.50}\color{red}\textsubscript{(0.26$\uparrow$)}                             & 32.63\color{ForestGreen}\textsubscript{(0.63$\downarrow$)}                            & \textbf{82.29}\color{red}\textsubscript{(4.75$\uparrow$)}                    & \textbf{19.96}\color{red}\textsubscript{(4.31$\uparrow$)}                          & \textbf{40.60}\color{red}\textsubscript{(2.18$\uparrow$)}                          \\ \bottomrule
\end{tabular}
}
\label{table1-rl-result1}
\end{table*}

\begin{table*}[!t]\small
\centering
\caption{Comparison of \our with other reward models on the DeepCoder dataset (\texttt{Avg@8}). \textbf{Bold} indicates the best results, \underline{Underline} denotes the second-best results. {$\uparrow(\downarrow)$} indicates the improvement (degradation) compared to the RLVR baseline.}
\label{table-rm-comparison}
\resizebox{\linewidth}{!}{%
\begin{tabular}{@{}llllll@{}}
\toprule
\multicolumn{1}{l}{\multirow{2}{*}{\textbf{\textsc{Models}}}} & \multicolumn{4}{c}{\textbf{\textsc{Benchmarks}}\texttt{(Avg@8)}}                                                                                                       & \multirow{2}{*}{\textbf{\textsc{Avg.}}} \\
\multicolumn{1}{c}{}                                & \multicolumn{1}{l}{LiveCodeBench} & \multicolumn{1}{l}{CodeContests} & \multicolumn{1}{l}{MBPP} & \multicolumn{1}{l}{CodeForces} &                               \\ \midrule
\multicolumn{6}{c}{\textsc{DeepCoder}}                                                                                                                                                                                    \\ \midrule
\textsc{Qwen3-8B-Base}                                       & 13.75                             & 19.03                            & 61.70                     & 5.35                           & 24.96                          \\
\quad\textit{w/ RLVR}                                    & \underline{23.60}                              & \underline{32.00}                              & \textbf{76.01}                  & \underline{16.43}                        & \underline{37.01}                            \\
\quad\textit{w/ \textsc{SkyworkRM}}                               & 18.50                              & 23.22                       & 67.59                      & 8.00                           & 29.33\color{ForestGreen}\textsubscript{(7.68$\downarrow$)}                      \\
\quad\textit{w/ \textsc{AceCodeRM}}                               & 22.75                              & 28.34                       & 71.94                      & 9.74                          & 33.19\color{ForestGreen}\textsubscript{(3.82$\downarrow$)}                      \\
\quad\textit{w/ \our (Ours)}                                  & \textbf{24.80}\color{red}\textsubscript{(1.20$\uparrow$)}                               & \textbf{33.94}\color{red}\textsubscript{(1.94$\uparrow$)}                             & \underline{75.90}\color{ForestGreen}\textsubscript{(0.11$\downarrow$)}                    & \textbf{19.61}\color{red}\textsubscript{(3.27$\uparrow$)}                          & \textbf{38.56}\color{red}\textsubscript{(1.55$\uparrow$)}                          \\ \midrule
\textsc{Qwen3-14B-Base }                                     & 21.37                             & 26.25                            & 71.09                    & 8.45                           & 31.79                            \\
\quad\textit{w/ RLVR}                                    & 25.94                            & \underline{34.30}                             & 75.45                   & \underline{20.79}                          & \underline{39.12}                        \\
\quad\textit{w/ \textsc{SkyworkRM}}                               & 24.05                   & 31.64                  & 74.26               & 10.35                 & 35.08\color{ForestGreen}\textsubscript{(4.04$\downarrow$)}                     \\
\quad\textit{w/ \textsc{AceCodeRM}}                               & \underline{25.98}                              & 32.11                       & \underline{81.22}                      & 13.35                          & 38.17\color{ForestGreen}\textsubscript{(0.95$\downarrow$)}                      \\
\quad\textit{w/ \our (Ours)}                                  & \textbf{27.55}\color{red}\textsubscript{(1.61$\uparrow$)}                             & \textbf{39.33}\color{red}\textsubscript{(5.03$\uparrow$)}                            & \textbf{81.61}\color{red}\textsubscript{(6.16$\uparrow$)}                    & \textbf{24.89}\color{red}\textsubscript{(4.10$\uparrow$)}                         & \textbf{43.35}\color{red}\textsubscript{(4.23$\uparrow$)}                           \\ \bottomrule
\end{tabular}
}
\end{table*}

\subsection{\our for RL Training}

\noindent\textbf{Setup} We use GRPO \citep{DBLP:journals/corr/abs-2501-12948} as our RL algorithm and implement it with VeRL\footnote{\href{https://github.com/volcengine/verl}{https://github.com/volcengine/verl}} \citep{DBLP:conf/eurosys/ShengZYWZZPL025}. We use \textsc{Qwen3-8B-Base} and \textsc{Qwen3-14B-Base} as our base models. Please refer to \autoref{rl-training-configuration} for training configuration.  We evaluate on four popular code generation benchmarks: LiveCodeBench \citep{DBLP:conf/iclr/JainHGLYZWSSS25}, CodeContests \citep{li2022competition}, MBPP \citep{DBLP:journals/corr/abs-2108-07732} and CodeForces \citep{penedo2025codeforces}. We report \texttt{Avg@8}, the average execution correctness over 8 independent generated solutions per problem, and set temperature to $0.6$ during evaluation across all experiments. Please refer to \autoref{rl-evaluation-datasets} for dataset details.

\subsubsection{High-Quality Supervision as Reward Signal}\label{results-rl-supervision}\label{results-deepcoder}\label{result-kodcode}

We evaluate the effectiveness of \our as a reward signal for RL training across three representative datasets: (1)~DeepCoder~\citep{deepcoder2025}, which consists of 24K high-quality problems with verified test cases, representing an ideal scenario for RLVR; (2)~KodCode~\citep{DBLP:conf/acl/XuLYZP25} and (3)~rStarCoder~\citep{DBLP:journals/corr/abs-2505-21297}, both of which are large-scale synthetic datasets where test cases are generated and verified automatically (see \autoref{rl-training-configuration} for details).

\noindent\textbf{Baselines.} We compare \our against RLVR, where rewards are derived from binary execution feedback on test cases. On the DeepCoder dataset, we additionally compare with two existing reward models: \textsc{SkyworkRM}, a general-purpose reward model~\citep{DBLP:journals/corr/abs-2507-01352}, and \textsc{AceCodeRM}, a code-specific reward model~\citep{DBLP:conf/acl/ZengJ0NCC25}.

\noindent\textbf{Results.}  As shown in \autoref{table1-rl-result1}, \our consistently matches or outperforms RLVR across all four datasets. On DeepCoder, where RLVR benefits from high-quality verified test cases, \our still achieves an average improvement of 1.55$\uparrow$ for \textsc{Qwen3-8B-Base} and even 4.23$\uparrow$ for \textsc{Qwen3-14B-Base}, with substantial gains on CodeContests~(34.3~$\to$~39.33) and CodeForces~(20.79~$\to$~24.89). On synthetic datasets, \our similarly outperforms RLVR, achieving 1.72$\uparrow$ on KodCode and 0.64$\uparrow$ on rStarCoder for the 8B model, and 1.71$\uparrow$ and 2.18$\uparrow$ respectively for the 14B model. We attribute these consistent gains to the dense nature of the reward signals provided by \our, which provides a richer learning gradient, allowing the policy to better capture structural nuances and explore a more diverse solution space.

Furthermore, as shown in \autoref{table-rm-comparison}, we compare \our against other reward models on DeepCoder. Both \textsc{SkyworkRM} and \textsc{AceCodeRM} fail to yield competitive gains during RL training, resulting in substantial degradation compared to RLVR (7.68$\downarrow$ and 3.82$\downarrow$ on average for 8B, respectively). In contrast, \our is the only reward model that surpasses RLVR. We attribute this to two factors: (1)~\our is trained on high-quality preference data derived from verified code problems; and (2)~the syntax-aware code extraction and validity-preserving reward shaping introduced in our method effectively stabilize the RL training process. To better understand the contribution of each component, we conduct a series of ablation studies in \autoref{ablation}.

\subsubsection{Scaling RL Training without Test Cases}\label{result-synthetic}

\begin{table}[H]
\centering
\small
\caption{Evaluation results (\texttt{Avg@8}) with scaled training on DeepCoder + synthetic data (44K problems). {$\uparrow(\downarrow)$} indicates the improvement (degradation) compared to the base model.}
\label{table3-rl-result3}
\resizebox{\linewidth}{!}{%
\begin{tabular}{@{}llllll@{}}
\toprule
\multicolumn{1}{l}{\multirow{2}{*}{\textbf{\textsc{Models}}}} & \multicolumn{4}{c}{\textbf{\textsc{Benchmarks}}\texttt{(Avg@8)}}                                                                                                       & \multirow{2}{*}{\textbf{\textsc{Avg.}}} \\
\multicolumn{1}{c}{}                                & \multicolumn{1}{l}{LiveCodeBench} & \multicolumn{1}{l}{CodeContests} & \multicolumn{1}{l}{MBPP} & \multicolumn{1}{l}{CodeForces} &                               \\ \midrule
\textsc{Qwen3-8B-Base}                                       & 13.75                             & 19.03                            & 61.70                     & 5.35                           & 24.96                          \\
\quad\textit{w/ \textsc{SkyworkRM}$^\dagger$}                               & 17.69                              & 23.48                       & 75.22                      & 7.52                           & 30.98                     \\
\quad\textit{w/ \textsc{AceCodeRM}}                               & 18.81                              & 27.56                       & 74.15                      & 9.20                         & 32.43                      \\
\quad\textit{w/ \our (Ours)}                                  & \textbf{25.76}\color{red}\textsubscript{(12.01$\uparrow$)}                               & \textbf{33.63}\color{red}\textsubscript{(14.60$\uparrow$)}                             & \textbf{76.81}\color{red}\textsubscript{(15.11$\uparrow$)}                    & \textbf{22.18}\color{red}\textsubscript{(16.83$\uparrow$)}                          & \textbf{39.60}\color{red}\textsubscript{(14.64$\uparrow$)}                          \\ \bottomrule
\multicolumn{6}{l}{\footnotesize $^\dagger$ Best checkpoint before training collapse.} \\
\end{tabular}
}
\end{table}

A fundamental limitation of RLVR is its dependence on test cases, which constrains the scale and diversity of training data. \our removes this bottleneck entirely, enabling RL training to scale by incorporating large amounts of synthetic problems without execution feedback. Following the pipeline of MathScale~\citep{DBLP:conf/icml/TangZWW24} and QueST~\citep{DBLP:journals/corr/abs-2510-17715}, we use TACO as the seed dataset, extract concepts from it, construct a concept graph, and leverage an LLM to generate 20K new problems (see \autoref{rl-training-configuration} for details). We then combine these with the 24K DeepCoder problems to form a 44K mixed training dataset and train \textsc{Qwen3-8B-Base} from scratch using \our.

\noindent\textbf{Results.} As shown in \autoref{table3-rl-result3}, training \textsc{Qwen3-8B-Base} with \our on the scaled 44K dataset yields an average improvement of \textbf{14.64} points over the base model, with substantial gains across all benchmarks (e.g., LiveCodeBench 13.75~$\to$~25.76, CodeForces 5.35~$\to$~22.18). This result demonstrates that \our can effectively leverage synthetic data to achieve continued performance gains through data scaling. We further evaluate \textsc{SkyworkRM} and \textsc{AceCodeRM} as alternative reward models under the same setting. During training, we found that \textsc{SkyworkRM} suffers from training collapse, with response length exploding and reward scores dropping rapidly, and we report results from the best checkpoint before collapse. \textsc{AceCodeRM} avoids collapse but exhibits continuous performance degradation throughout training. In contrast, \our maintains stable training dynamics and achieves consistent improvements.

\subsection{\our for Test-Time Scaling}

\subsubsection{Comparison with Other TTS Methods}\label{comparison-tts}

\noindent\textbf{Baselines.} We compare with two types of TTS methods: (1) Unit Test TTS method, represented by CURE \citep{DBLP:journals/corr/abs-2506-03136}, which generates unit tests for each problem and selects the candidate solution that passes most of them. (2) Other reward models, where we use \textsc{SkyworkRM} and \textsc{AceCodeRM} as representatives of general-purpose and code-specific RMs.

\noindent\textbf{Evaluation Details.} For fair comparisons, we use \textsc{ReasonFlux-Coder-7B} trained with CURE as the policy model, since CURE requires the same model to generate unit tests for BoN sampling. We generate 8 candidate solutions using \textsc{ReasonFlux-Coder-7B} for each problem on four benchmarks. For CURE, we generate 8 unit tests per problem. We report \texttt{BoN@8} across all benchmarks and set temperature to 0.6 during evaluation. As in \autoref{Bon-result1}, \our substantially outperforms alternative reward models and achieves performance comparable to CURE, attains a favorable trade-off between effectiveness and efficiency. We define latency as the end-to-end wall-clock latency per problem, measured from the moment a prompt is provided to the policy model until the final selection is returned. For unit test TTS methods, the latency can be expressed as:

\begin{figure*}[t]
\centering
    \includegraphics[width=\linewidth]{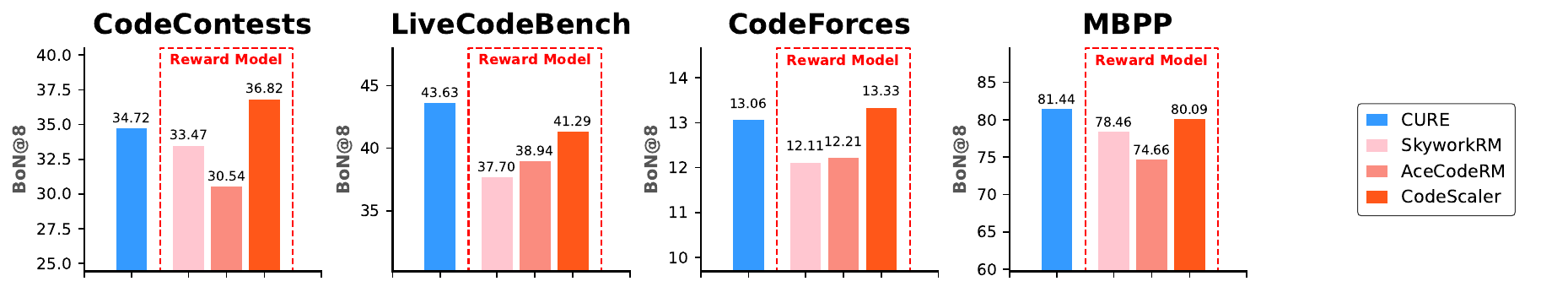}
    \caption{Comparison of Best-of-N (\texttt{BoN@8}) performance across four code generation benchmarks using different test-time scaling methods. \our consistently outperforms other reward models and achieves performance comparable to CURE.}
    \label{Bon-result1}
\end{figure*}

\begin{equation}
    \small
    \label{eq:cure_latency}
    \scalebox{0.95}{$
    \mathcal{L}_{\textsc{UnitTest}} =
    \underbrace{N \cdot T_{\text{gen}}}_{\text{generating candidates}}
    +
    \underbrace{M \cdot T_{\text{test-gen}}}_{\text{generating unit tests}}
    +
    \underbrace{N \cdot M \cdot T_{\text{exec}}}_{\text{sandbox execution}}
    $},
\end{equation}
where $N$ denotes the number of candidate solutions and $M$ denotes the number of generated unit tests.

Reward model TTS assigns scalar scores to candidate solutions without requiring test generation or code execution. The latency of \our can be expressed as:
\begin{equation}
    \label{eq:rm_latency}
    \scalebox{0.95}{$
    \mathcal{L}_{\textsc{CodeScaler}} =
    \underbrace{N \cdot T_{\text{gen}}}_{\text{generating candidates}}
    +
    \underbrace{N \cdot T_{\text{RM}}}_{\text{reward model scoring}}$
    },
\end{equation}
where $T_{\text{RM}}$ denotes the inference time of the reward model for a single candidate.

\begin{wraptable}{r}{0.44\linewidth}
\centering
\small
\caption{Latency comparison between CURE and \our.}
\label{tab:latency-evaluation}
\resizebox{\linewidth}{!}{%
\begin{tabular}{@{}lcc@{}}
\toprule
\textsc{Time(s)} & \textsc{CURE} & \our \\ \midrule
\textsc{Unit Test Gen.} & 979.3 & N/A \\
\textsc{Execution} & 516.7 & N/A \\
\textsc{RM Compute} & N/A & 146.1 \\ \midrule
\textsc{Avg(s/question).} & 3.20 & \textbf{0.31} \\ \bottomrule
\end{tabular}
}
\end{wraptable}

Since candidate generation is shared across both methods, we omit the $N \cdot T_{\text{gen}}$ term when comparing latency. We evaluate \textsc{CURE} and \our as representative unit test TTS and reward model TTS methods, respectively. All latency measurements are conducted on the CodeForces (467 questions) using a single NVIDIA A100 GPU with vLLM backbone. We report results under the \texttt{BoN@8} setting, where 8 candidate solutions are generated per problem, and for \textsc{CURE}, 8 unit tests are generated and executed for each question. As shown in \autoref{tab:latency-evaluation}, \our achieves an approximately 10× speedup over \textsc{CURE}. \our scores candidates using only a single forward pass. This substantially reduces computational overhead and resource consumption, making \our particularly well suited for real-world deployment.

\subsubsection{Cross-Domain Transfer of \our}\label{results-rmbench}

To further assess the model, we evaluated \our on standard reward model benchmarks. Although \our was not designed as a general purpose reward model, testing it on these benchmarks still provides valuable insights. We used RM-Bench \citep{DBLP:conf/iclr/Liu0M00L25} to compare \our against \textsc{SkyworkRM} and \textsc{AceCodeRM}, as shown in \autoref{tab:rmbench}. Beyond the clear improvement in code tasks (73.6 → 76.9), we found a surprising result: even though \our was trained only on code, it also improved in general (Chat, 80.6 → 83.0) and reasoning domains (Math, 75.0 → 79.9).  Previous research in SFT and RL has shown that including math or code data can boost a model's general abilities \citep{DBLP:journals/tmlr/PettySL25,ma2023training,li2025can, DBLP:journals/corr/abs-2507-00432}. Our results indicate that this also applies to reward model training. Learning to judge code sharpens the model's overall judgment even outside of the code domain.

\begin{table*}[t]\small
\caption{Evaluation results on RM-Bench across diverse reward models. \textbf{Bold} indicates the best results. {$\uparrow(\downarrow)$} indicates the improvement (degradation) compared to the SkyworkRM-8B.}
\label{tab:rmbench}
\centering
\resizebox{\linewidth}{!}{%
\begin{tabular}{@{}lllllllll@{}}
\toprule
\multicolumn{1}{l}{\multirow{2}{*}{\textbf{\textsc{Models}}}} & \multicolumn{7}{c}{\textbf{\textsc{RM-Bench}}}                           & \multirow{2}{*}{\textbf{\textsc{Avg.}}} \\
\multicolumn{1}{c}{}                                        & Chat          & Math  & Code  & Safety & Hard  & Normal & Easy  &                               \\ \midrule

\textsc{SkyworkRM-8B}                                  & \underline{80.6}         & \underline{75.0}    & \underline{73.6} & \textbf{96.5}  & \underline{67.0} & \underline{85.5}  & \underline{91.8} & \underline{81.4}                         \\
\textsc{AceCodeRM-7B}                                                & 66.7 & 65.3  & 66.9  & 89.9   & 62.2  & 74.4   & 79.9  & 72.2                          \\
\textsc{AceCodeRM-32B}                                               & 73.7          & 70.5  & 72.1  & 88.0     & 65.5  & 78.3   & 78.3  & 76.1                          \\
\textsc{CodeScaler-8B} \textit{(Ours)}                                                 & \textbf{83.0}\color{red}\textsubscript{(2.4$\uparrow$)}         & \textbf{79.9}\color{red}\textsubscript{(4.9$\uparrow$)}  & \textbf{76.9}\color{red}\textsubscript{(3.3$\uparrow$)}  & \underline{96.4}\color{ForestGreen}\textsubscript{(0.1$\downarrow$)}  & \textbf{71.8}\color{red}\textsubscript{(4.8$\uparrow$)} & \textbf{87.9}\color{red}\textsubscript{(2.4$\uparrow$)}  & \textbf{92.5}\color{red}\textsubscript{(0.7$\uparrow$)}  & \textbf{84.1}\color{red}\textsubscript{(2.7$\uparrow$)}                          \\
\bottomrule
\end{tabular}
}
\end{table*}

\subsubsection{Multi-Lingual Transfer of \our}

We evaluate \our on HumanEval-X \citep{zheng2023codegeex}, which spans five programming languages: Python, C++, Java, JavaScript, and Go. As shown in \autoref{tab:multilingual-bon}, \our consistently matches or outperforms \textsc{SkyworkRM} across all languages. We attribute this generalization to the nature of the reward signal learned by our model. Rather than relying on language-specific syntax, \our implicitly models functional correctness, capturing properties such as correct handling of edge cases and logically sound reasoning. These properties are inherently language-agnostic, as a correct algorithmic solution in Python shares the same underlying structure as its counterpart in C++ or Java.

\begin{table}[H]
\centering
\caption{Multi-lingual \texttt{BoN@8} evaluation on HumanEval-X using \textsc{Qwen3-8B} as the policy model. {$\uparrow(\downarrow)$} indicates the improvement (degradation) compared to the Pass@1.}
\label{tab:multilingual-bon}
\resizebox{0.75\linewidth}{!}{%
\begin{tabular}{lccc}
\toprule
\textbf{Language} & \textbf{Pass@1} & \textbf{SkyworkRM BoN@8} & \textbf{\our BoN@8} \\
\midrule
Python     & 30.7 & 34.1 & \textbf{34.8}\color{red}\textsubscript{(4.1$\uparrow$)}  \\
C++        & 28.0 & 32.9 & \textbf{33.2}\color{red}\textsubscript{(5.2$\uparrow$)}  \\
Java       & 26.8 & 32.9 & \textbf{34.8}\color{red}\textsubscript{(8.0$\uparrow$)}  \\
JavaScript & 24.4 & 25.6 & \textbf{25.6}\color{red}\textsubscript{(1.2$\uparrow$)}  \\
Go         & 21.4 & 24.4 & \textbf{25.0}\color{red}\textsubscript{(3.6$\uparrow$)}  \\
\bottomrule
\end{tabular}
}
\end{table}


\section{Ablation and Analysis}\label{ablation}

\subsection{Reward Quality of \our}

To validate that the reward signals produced by \our faithfully reflect code correctness, we analyze training trajectory records from \textsc{Qwen3-8B-Base} on DeepCoder, where ground-truth execution labels are available. We evaluate \our from three complementary perspectives:

\textbf{Pairwise Ranking Accuracy.} Given a randomly sampled pair of one correct and one incorrect solution, \our assigns a higher score to the correct one 87.9\% of the time, corresponding to an AUROC of 0.879. This indicates that \our possesses strong discriminative ability to distinguish correct solutions from incorrect ones.

\textbf{Score Separation.} The mean RM score for correct solutions is 10.94, compared to 3.77 for incorrect ones, yielding a Cohen's $d$ of 1.71. This large effect size confirms that \our produces well-separated score distributions, capturing meaningful quality differences rather than relying on marginal distinctions.

\textbf{Group-Level Selection Accuracy.} Within each rollout group, the candidate receiving the highest RM score is correct approximately 80\% of the time, achieving a Top-1 selection accuracy of 79.8\%. This directly validates the practical utility of \our for Best-of-N selection, where the top-ranked candidate is chosen as the final output.

These results demonstrate that \our provides reliable and well-calibrated reward signals, supporting its dual role as both a training reward for RL optimization and a test-time ranker for Best-of-N sampling.





\subsection{Ablation on \our Training Data}

In reward model training, we use the DeepCoder training dataset, which provides high-quality, verified test cases and thus offers reliable supervision for reward learning. To study the impact of training data quality, we conduct a controlled comparison by collecting RLVR trajectories from \textsc{Qwen3-8B-Base} trained on KodCode and using them to train an alternative reward model. Since KodCode relies on synthesized test cases, it can be viewed as a lower-quality variant in terms of supervision. We then use these two reward models to train \textsc{Qwen3-8B-Base} on the rStarCoder dataset. 

From the \autoref{tab:codescaler-variant-data}, we observe that \our trained on DeepCoder consistently outperforms its counterpart. In particular, the model trained with DeepCoder rollouts achieves a 0.71$\uparrow$ on LiveCodeBench (23.79 → 24.5), 2.56$\uparrow$ on CodeContests (28.19 → 30.75) and 4.37$\uparrow$ on CodeForces (11.58 → 15.95). These findings highlight the critical role of high-quality, verified training data in reward model learning and further demonstrate the performance gap between verified code datasets and synthetic datasets.


\begin{table}[!ht]
\centering
\footnotesize
\caption{Ablation on \our training data. Results compare RL training performance using different trajectory sources.}
\label{tab:codescaler-variant-data}
\resizebox{0.6\linewidth}{!}{%
\begin{tabular}{@{}llll@{}}
\toprule
\multicolumn{1}{l}{\multirow{2}{*}{\textbf{\textsc{Traj.}}}} & \multicolumn{3}{c}{\textbf{\textsc{Benchmarks}}\texttt{(Avg@8)}}                                                                     \\
\multicolumn{1}{l}{}                       & \multicolumn{1}{c}{LiveCodeBench} & \multicolumn{1}{c}{CodeContests} & \multicolumn{1}{c}{CodeForces} \\ \midrule
\multicolumn{4}{c}{\textsc{rStarCoder}}                                                                                                                       \\ \midrule

\textsc{KodCode}                                        &
23.79                            &
28.19                        &
11.58                         \\

\textsc{DeepCoder}                                         &
\textbf{24.50}\color{red}\textsubscript{(0.71$\uparrow$)}                             &
\textbf{30.75}\color{red}\textsubscript{(2.56$\uparrow$)}                          &
\textbf{15.95}\color{red}\textsubscript{(4.37$\uparrow$)}                           \\ \bottomrule
\end{tabular}
}
\end{table}

\subsection{Ablation on \our Components}

\begin{figure}[t]
\centering
    \includegraphics[width=\linewidth]{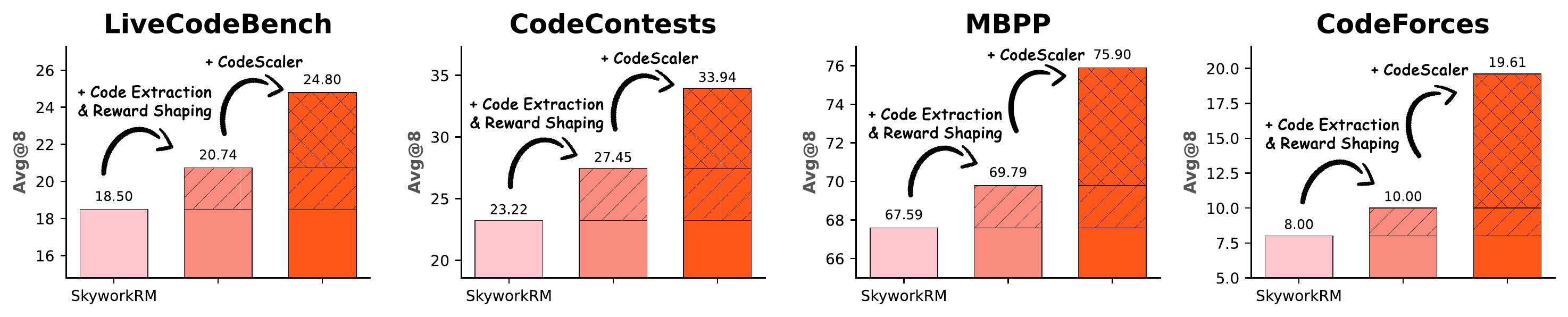}
    \caption{Ablation study on \our components. Results compare RL performance using different reward model variants.}
    \label{ablation-ast}
\end{figure}

To enable stable RL training, we make two key contributions. On the training side, we introduce syntax-aware code extraction and validity-preserving reward shaping to stabilize the reward signal. On the data side, we construct high-quality code preference pairs to enhance the reward model’s capability. In this section, we conduct ablation to validate the effectiveness of both components. We train \textsc{Qwen3-8B-Base} on DeepCoder under three settings: (1) the original \textsc{SkyworkRM}, (2) \textsc{SkyworkRM} with syntax-aware code extraction and validity-preserving reward shaping, and (3) \our with syntax-aware code extraction and validity-preserving reward shaping. Results see \autoref{ablation-ast}.

We observe a notable improvement when augmenting \textsc{SkyworkRM} with syntax-aware code extraction and validity-preserving reward shaping. In particular, performance improves by 2.24$\uparrow$ on LiveCodeBench (18.5 → 20.74), 4.23~$\uparrow$ on CodeContests (23.22 → 27.45) and 2.2~$\uparrow$ on MBPP (67.59 → 69.79), indicating that the proposed techniques effectively stabilize the reward signal and lead to better RL optimization. Replacing \textsc{SkyworkRM} with \our further yields substantial performance gains, with improvements of 4.06$\uparrow$ on LiveCodeBench (20.74 → 24.80), 6.49~$\uparrow$ on CodeContests (27.45 → 33.94), 6.11~$\uparrow$ on MBPP (69.79 → 75.90), and 9.61~$\uparrow$ on CodeForces (10.00 → 19.61), highlighting the superior reward quality of \our.

\subsection{Analysis on Pass Ratio Feedback}\label{analysis-binary}

\begin{table}[!ht]
\centering
\caption{Ablation study on preference pair construction. Results compare RL training performance using different \our variant. \textbf{Bold} indicates the best results, \underline{Underline} denotes the second-best results. {$\uparrow(\downarrow)$} indicates the improvement (degradation) compared to the RLVR (Binary) baseline.}
\label{tab:codescaler-variant-partial}
\resizebox{\linewidth}{!}{%
\begin{tabular}{@{}llllll@{}}
\toprule
\multicolumn{1}{l}{\multirow{2}{*}{\textbf{\textsc{Models}}}} & \multicolumn{4}{c}{\textbf{\textsc{Benchmarks}}\texttt{(Avg@8)}}                                                                                                       & \multirow{2}{*}{\textbf{\textsc{Avg.}}} \\
\multicolumn{1}{c}{}                                & \multicolumn{1}{l}{LiveCodeBench} & \multicolumn{1}{l}{CodeContests} & \multicolumn{1}{l}{MBPP} & \multicolumn{1}{l}{CodeForces} &                               \\ \midrule
\multicolumn{6}{c}{\textsc{DeepCoder}}                                                                                                                                                                                    \\ \midrule
\textsc{Qwen3-8B-Base}                                       &
13.75                             &
19.03                            &
61.70                     &
5.35                           &
24.96                          \\
\quad\textit{w/ RLVR (Binary)}                                    &
23.60                              &
32.00                              &
\textbf{76.01}                  &
16.43                        &
37.01                            \\

\quad\textit{w/ \our (Binary)}                                  &
\textbf{24.80}\color{red}\textsubscript{(1.20$\uparrow$)}                               & \underline{33.94}\color{red}\textsubscript{(1.94$\uparrow$)}                             & \underline{75.90}\color{ForestGreen}\textsubscript{(0.11$\downarrow$)}                    & \underline{19.61}\color{red}\textsubscript{(3.27$\uparrow$)}                          & \textbf{38.56}\color{red}\textsubscript{(1.55$\uparrow$)}                          \\

\quad\textit{w/ \our (Pass Ratio)}                                  & \underline{24.20}\color{red}\textsubscript{(0.60$\uparrow$)}                               & \textbf{35.04}\color{red}\textsubscript{(3.04$\uparrow$)}                             & 71.09\color{ForestGreen}\textsubscript{(4.92$\downarrow$)}                    &
\textbf{21.44}\color{red}\textsubscript{(5.01$\uparrow$)}                          & \underline{37.94}\color{red}\textsubscript{(0.93$\uparrow$)}                          \\

 \bottomrule
\end{tabular}
}
\end{table}

\begin{table}[!ht]
\small
\centering
\caption{Ablation study on preference pair construction. Results compare Best-of-N sampling performance using different \our variant. \textbf{Bold} indicates the best results.}
\label{tab:codescaler-variant-partial-bon}
\resizebox{\linewidth}{!}{%
\begin{tabular}{@{}llllll@{}}
\toprule
\multicolumn{1}{l}{\multirow{2}{*}{\textbf{\textsc{RM Variants}}}} & \multicolumn{4}{c}{\textbf{\textsc{Benchmarks}}\texttt{(BoN@8)}}                                                                                                       & \multirow{2}{*}{\textbf{\textsc{Avg.}}} \\
\multicolumn{1}{c}{}                                & \multicolumn{1}{l}{LiveCodeBench} & \multicolumn{1}{l}{CodeContests} & \multicolumn{1}{l}{MBPP} & \multicolumn{1}{l}{CodeForces} &                                                                                                                                                                                                      \\ \midrule

\our \textit{(Binary)}                                  &
\textbf{41.29}                               &
\textbf{36.82}                             &
\textbf{80.09}                   &
\textbf{13.33}                     &
\textbf{42.88}                    \\

\our \textit{(Pass Ratio)}                                  &
40.84                          &
35.88                            &
79.64                &
13.28                        &
42.41                  \\

 \bottomrule
\end{tabular}
}
\end{table}

Code generation problems are typically formulated with binary feedback, where a solution either passes all test cases or fails. However, a finer-grained supervision signal can be obtained from the test case pass ratio, which measures the proportion of test cases a solution successfully passes. This distinction also reflects real-world contest settings: OI-style contests usually assign scores per test case, whereas ICPC-style contests adopt a strict pass/fail criterion.

In this section, we investigate whether incorporating pass-ratio supervision is beneficial during reward model training.

\subsubsection{Impact on Reward Model Training}

When constructing preference pairs, we adopt a strict binary construction where solutions that pass all test cases are treated as positive samples, while those that fail any test case are treated as negative. In contrast, prior work \citep{DBLP:conf/acl/ZengJ0NCC25} employs a looser criterion, labeling solutions with a pass ratio greater than a threshold as positives. This design choice suggests that reward models may potentially learn from partially correct solutions.

To evaluate whether such threshold-based construction is necessary, we train a variant of \our where for each problem, solutions with a pass ratio above 0.7 are treated as positive samples, even if they are not fully correct.

As shown in \autoref{tab:codescaler-variant-partial} and \autoref{tab:codescaler-variant-partial-bon}, our binary \our outperforms the pass ratio counterpart in both RL training and Best-of-N sampling. We attribute this degradation to reduced label validity. Unlike binary positives, which correspond to fully verified solutions, high pass-ratio solutions may still contain latent logical errors due to incomplete test coverage. Treating such partially correct solutions as positives introduces noise into preference supervision, weakening the reward model’s ability to reliably distinguish code quality. Our results indicate that the reward model can still learn sufficiently fine-grained signals from strictly valid binary supervision.

\subsection{\our at Different Scales}
Reward models are often used as lightweight plugins in downstream tasks. Therefore, it is important to provide models of various sizes to meet different resource constraints. We trained different sizes of \our from 1.7B, 4B, to 8B parameters to meet this need. We evaluated these models in RL training, BoN task and RM-Bench. As shown in \autoref{tab:codescaler-variant-parameters}, performance improves consistently as the model size increases. Our smallest model, \our-1.7B, already performs comparably to \textsc{AceCodeRM}-7B in RL training. Furthermore, \our-4B surpasses SkyworkRM-8B on RM-Bench. See \autoref{appdix:rm-bench} for full results on RM-Bench.
\begin{table}[!ht]
\small
\centering
\caption{Ablation study on \our in diverse scales. Evaluation across code benchmarks, BoN sampling and RM-Bench. {$\uparrow(\downarrow)$} indicates the improvement (degradation) compared to the SkyworkRM-8B.}
\label{tab:codescaler-variant-parameters}
\resizebox{0.75\linewidth}{!}{%
\begin{tabular}{llll}
\toprule
\multicolumn{1}{l}{\multirow{2}{*}{\textbf{\textsc{RM Variants}}}} & \multicolumn{1}{c}{\textbf{\textsc{Benchmarks}}} &\multicolumn{1}{c}{\textbf{\texttt{BoN@8}}}   &\multicolumn{1}{c}{\textbf{\textsc{RM-Bench}}}                                                                 \\
\multicolumn{1}{l}{}                       & \multicolumn{1}{c}{\textsc{Avg.}}  & \multicolumn{1}{c}{\textsc{Avg.}} & \multicolumn{1}{c}{\textsc{Avg.}}\\

\midrule
\textsc{SkyworkRM}-8B & 29.33 &41.17 & 81.46\\
\textsc{AceCodeRM}-7B & 33.19\color{red}\textsubscript{(3.86$\uparrow$)} &40.48\color{ForestGreen}\textsubscript{(0.69$\downarrow$)} &72.20\color{ForestGreen}\textsubscript{(9.26$\downarrow$)}\\ \midrule

\our-1.7B  & 35.40\color{red}\textsubscript{(6.07$\uparrow$)}  & 38.16\color{ForestGreen}\textsubscript{(3.01$\downarrow$)}  & 78.83\color{ForestGreen}\textsubscript{(2.63$\downarrow$)} \\
\our-4B  & 37.20\color{red}\textsubscript{(7.87$\uparrow$)}  & 39.82\color{ForestGreen}\textsubscript{(1.35$\downarrow$)}  & 82.86\color{red}\textsubscript{(1.40$\uparrow$)} \\
\our-8B  & \textbf{38.56}\color{red}\textsubscript{(9.23$\uparrow$)}  & \textbf{44.15}\color{red}\textsubscript{(3.10$\uparrow$)}  &\textbf{84.05}\color{red}\textsubscript{(2.59$\uparrow$)}  \\
\bottomrule
\end{tabular}
}
\end{table}



\section{Related Work}

\noindent\textbf{LLMs for Code Generation.} LLMs are recently studied for code generation based on their outstanding capability \citep{deepcoder2025}. Early milestones established that transformer-based LLMs can synthesize functional programs across a wide range of tasks, sparking extensive follow-up research \citep{DBLP:journals/corr/abs-2107-03374, DBLP:conf/emnlp/0034WJH21}. Subsequent models and open-source efforts have further improved performance through architectural scaling, curated pretraining corpora, and supervised instruction fine-tuning \citep{DBLP:conf/acl/ZengJ0NCC25, DBLP:journals/corr/abs-2305-07922, lozhkov2024starcoder2stackv2, rozière2024codellamaopenfoundation, guo2024deepseekcoderlargelanguagemodel, huang2024swiftcoder}. Beyond supervised fine-tuning,  in-context learning enables LLMs to adapt to novel coding tasks or adhere to strict output formats using demonstrations and prompts \citep{chen2023programthoughtspromptingdisentangling, li2023acecoderutilizingexistingcode, DBLP:conf/naacl/PatelRBD24, huang2024effilearner}. Some works are involved with training models with distilled trajectories from advanced models \citep{DBLP:journals/corr/abs-2504-01943,openr1,wu2026largescaleterminalagentictrajectory}. More recently, RL has been widely adopted to improve LLM's generation ability further \citep{deepcoder2025, DBLP:journals/corr/abs-2510-17715}, where the test cases evaluation is essential to obtain the reliable reward signals \citep{DBLP:journals/tmlr/ShojaeeJTR23, DBLP:journals/corr/abs-2402-01391, DBLP:conf/icml/GehringZCMCS25}.  

\noindent\textbf{Datasets for Code Generation.} Research in code generation has relied on a spectrum of datasets ranging from real human-written benchmarks to large-scale synthetic data tailored for training. Classic benchmarks derived from real software repositories and programming problems provide canonical tasks with unit tests grounded in human logic and correctness criteria~\citep{DBLP:journals/corr/abs-2107-03374, dai2024mhpp, DBLP:conf/iclr/JainHGLYZWSSS25, qing2025effibench}. In addition to these benchmarks, real code corpora such as GitHub-sourced files and curated collections (e.g., CodeSearchNet \citep{husain2020codesearchnetchallengeevaluatingstate} or class-level datasets collected from open-source projects) offer large paired examples of code and documentation for both pretraining and fine-tuning \citep{rahman2025largescaleclasslevelbenchmarkdataset}. While verified test cases provide reliable supervision, they are costly and limited in scale. CodeContests \citep{li2022competition} contains 13K problems, and TACO \citep{DBLP:journals/corr/abs-2312-14852} provides only 25K training instances. To address scarcity and improve diversity, recent work has explored synthetic dataset generation using LLMs. Datasets such as OpenCodeInstruct generate millions of instruction–solution pairs with test cases and automatic quality assessments to enable robust supervised fine-tuning \citep{ahmad2025opencodeinstructlargescaleinstructiontuning}. Other synthetic pipelines like UnitCoder~\citep{ma2025unitcoderscalableiterativecode} leverage model-generated unit tests to guide and validate large synthetic corpora of verifiable programs, demonstrating improvements in both breadth and functional correctness \citep{ma2025unitcoderscalableiterativecode}. More recent synthetic training data like KodCode \citep{DBLP:conf/acl/XuLYZP25}, rStarCoder \citep{DBLP:journals/corr/abs-2505-21297}, and QueST~\citep{DBLP:journals/corr/abs-2510-17715} systematically generate diverse question–solution–test triplets with verification, enabling both supervised fine-tuning and execution-aware training for modern code LLMs.

\noindent\textbf{Reward Models from General to Domain-Specific.}  Reward models have become a cornerstone of modern alignment for LLMs, serving as learned proxies that approximate human qualitative judgments and guide reinforcement learning optimization when explicit task rewards are infeasible to define manually. In the general reinforcement learning from human feedback (RLHF) paradigm, a reward model is typically trained on human preference comparisons to assign scalar values to model outputs, thereby capturing human intent and enabling policy models to be optimized via reinforcement methods such as proximal policy optimization (PPO) \citep{ouyang2022traininglanguagemodelsfollow, schulman2017proximalpolicyoptimizationalgorithms}. More recent research has explored enhancements to reward model training and evaluation, including sequence-to-sequence reward modeling to capture richer language feedback \citep{zhou2025sequencesequencerewardmodeling}. In the specific domain of code generation, developing effective reward models is particularly challenging due to the need to quantitatively assess functional correctness and implementation quality. Recent work on BigCodeArena highlights the limitations of generic preference-based RMs by collecting extensive human evaluations under execution settings and proposing benchmarks that better reflect real coding preferences and execution outcomes \citep{zhuo2025bigcodearenaunveilingreliablehuman}. To reduce reliance on execution-based supervision, several works have explored learning reward models from code data \citep{DBLP:journals/corr/abs-2410-02229, DBLP:conf/acl/ZengJ0NCC25}. However, most evaluations focus on Best-of-N (BoN) sampling or reward model benchmarks, rather than using reward models as the primary signal for reinforcement learning. Empirically, RM-based RL has not been shown to outperform rule-based RLVR, leaving a clear performance gap between learned rewards and rule-based verification \citep{DBLP:conf/acl/ZengJ0NCC25}.

\section{Conclusion}

We introduce \our, a code reward model designed for both RL training and test-time inference, along with syntax-aware code extraction and validity-preserving reward shaping to stabilize reward signals. Experimental results show that \our consistently outperforms execution-based rewards during RL training. Scaling to 44K problems with additional synthetic data, \our yields +14.64 points improvement over the base model without requiring any test cases. At inference time, \our achieves performance comparable to unit test TTS methods with 10$\times$ lower latency. It also surpasses existing reward models on RM-Bench in code and other domains. We hope this work will stimulate broader interest in code reward models.


\clearpage 

\bibliography{references}
\bibliographystyle{abbrvnat}

\clearpage

\newpage
\appendix
\section*{Appendix}

\section{Broader Impact}
\label{appdix:impact}
This work introduces a reward model designed to enhance the code generation capabilities of LLMs. By providing dense scalar feedback and mitigating reward hacking, our approach facilitates the development of more reliable and functional software assistants. These advancements have the potential to democratize programming, increase developer productivity, and accelerate innovation in fields relying on complex algorithmic problem-solving. Furthermore, our findings on test-time scaling offer a computationally efficient alternative to traditional unit-test-based verification, potentially reducing the energy consumption and environmental footprint associated with large-scale inference.

Despite these benefits, the deployment of enhanced code LLMs carries inherent security and safety risks. More capable code generation models could inadvertently introduce subtle security vulnerabilities into software or be misused to automate the creation of malicious exploits. Additionally, while our method reduces the reliance on expensive human-curated data, it may inherit or amplify biases present in the synthetic or on-policy training sets. We encourage practitioners to utilize our provided model documentation and licenses to ensure responsible deployment and to maintain human oversight in critical software development pipelines.

\section{RM Training Details}\label{appendix:rm-training-details}
\noindent\textbf{Preference Construction Details.} We train \textsc{Qwen3-8B-Base} on the DeepCoder dataset using GRPO for 250 steps. During training, we set the temperature to 0.6 and top-$p$ to 0.95. For each problem, we rollout 8 candidate solutions and construct preference pairs by treating solutions that pass all test cases as positive and those that fail any test case as negative. We discard rollouts from the first 100 steps, resulting in a total of 52,574 preference pairs.

\noindent\textbf{Data Augmentation.} To improve the robustness of our reward model, we augment the preference pairs by adding misaligned pairs \((\vq, \tilde{\vc})\), where \(\tilde{\vc}\) is a solution code generated for a randomly different problem. This encourages the reward model to capture semantic alignment between the problem specification and the generated code, to prevent potential hacks that LLMs generate irrelevant solution codes of the corresponding question. In practice, we augment $\sim30\%$ data samples and combine them with the original curated preference pairs for reward model training.

\noindent\textbf{RM Training Configuration.} We use Skywork-Reward-V2-Qwen3-8B as our reward model for training, which achieves state-of-the-art performances on multiple RM benchmarks. We train it on collected preference data for $3$ epochs by AdamW, with a learning rate of $1e-6$.

\section{RL Training Details}\label{rl-training-configuration}
\subsection{Policy Gradient Optimization}

We adopt Group Relative Policy Optimization (GRPO) as our policy gradient optimization method. GRPO estimates advantages using group-normalized rewards. Specifically, for a given question $x$, the policy generates a group of $G$ responses $\{r_1, r_2, \ldots, r_G\}$. We extract code $\{c_1, c_2, \ldots, c_G\}$ from these responses and assign a transformed reward $R^\prime_i$ to each using the reward model. The GRPO optimization objective is then defined as:
\begin{equation}
\begin{aligned}
\mathcal{J}_{\text{GRPO}}(\theta) =  \mathbb{E}_{\vx \sim \mathcal{D}, \vo \sim \pi_\theta} &\Bigg[ \sum_{t=1}^{|\vo|} \min \Big( r_{i,t}(\theta) \cdot A, \tilde r_{i,t}(\theta) \cdot A \Big) \Bigg]
 - \beta \cdot \mathcal{D}_{\text{KL}}[\pi_\theta \| \pi_{\text{ref}}]
\end{aligned}
\end{equation}

where $\tilde r_{i,t}(\theta)=\text{clip}(r_{i,t}(\theta), 1-\varepsilon, 1+\varepsilon)$, and $r_{i,t}(\theta) = \frac{\pi_\theta(\vo_{i,t} \mid \vx, \vo_{i,<t})}{\pi_{\theta_{\text{old}}}(\vo_{i,t} \mid \vx, \vo_{i,<t})}$ denotes the importance sampling ratio. $A = \frac{R^\prime_i - \text{mean}(\{R^\prime_1,R^\prime_2,\cdots,R^\prime_G)}{\text{std}(\{R^\prime_1,R^\prime_2,\cdots,R^\prime_G\}}$ means the group-normalized advantage estimate. The KL divergence term prevents the policy from deviating too far from the reference model. Although the KL term has been omitted in some recent approaches, we observe that keeping KL regularization is beneficial for stabilizing reward model based RL training, especially when the quality of synthetic training data is not strictly controlled.

\subsection{RL Training Datasets}
\subsubsection{DeepCoder}

We use the DeepCoder training dataset for both reward model training and reinforcement learning. The dataset consists of three sources:
(1) verified problems from TACO,
(2) verified problems from PrimeIntellect's SYNTHETIC-1 dataset, and
(3) LiveCodeBench problems submitted between May 1, 2023 and July 31, 2024.

To ensure high data quality, DeepCoder applies a series of strict filtering criteria. First, each problem is automatically verified using an external official solution, and only problems whose official solutions pass all unit tests are retained. Second, each problem is required to include at least five unit tests. Third, duplicate problems are removed to prevent data leakage. In addition, the dataset is curated to ensure no overlap with the LiveCodeBench test set (08/01/24-02/01/25).

\subsubsection{KodCode \& rStarCoder}
In \autoref{result-kodcode}, we use KodCode and rStarCoder as representative examples of synthetic datasets.

KodCode employs a self-verification pipeline to validate automatically generated tests. Specifically, \textit{GPT-4o-0513} is used to generate both candidate solutions and corresponding test cases, and the generated solutions are executed against these tests to verify correctness. In addition, KodCode enforces 100\% branch coverage using \texttt{pytest-cov}. However, branch coverage serves only as a proxy for test quality and does not explicitly guarantee completeness of the generated test cases. During training, we use a dataset from \href{https://huggingface.co/datasets/KodCode/KodCode-V1-SFT-R1}{https://huggingface.co/datasets/KodCode/KodCode-V1-SFT-R1} and select only problems labelled with `online judge' style.

rStarCoder prompts \textit{GPT-4o} to generate two functions for each problem: one for test input generation and another for input validation. It then applies a mutual verification procedure to construct synthetic problems. Specifically, \textit{QWQ-32B} is used to generate 16 long-reasoning candidate solutions per problem, which are executed on at least 50 generated test inputs. If a majority of these candidate solutions produce identical outputs across the entire test set, both the consistent output set and the solutions that generate it are treated as correct. Compared to KodCode, this verification pipeline is more robust. Nevertheless, it still inherits limitations related to test completeness: challenging or adversarial test cases are often unlikely to pass the mutual verification process and are therefore discarded. During training, we use dataset from \href{https://huggingface.co/datasets/microsoft/rStar-Coder}{https://huggingface.co/datasets/microsoft/rStar-Coder} and select synthetic\_rl\_testcase split.

\subsubsection{Synthetic Problems}

In \autoref{result-synthetic}, we follow QueST to generate new code problems, using TACO as the seed dataset. For each problem $\vq \in \mathbf{Q}_{\text{seed}}$, we prompt \textit{GPT-4o} to extract a set of underlying concepts, including both topics and knowledge points.

Two concepts are considered to form a reasonable combination if they frequently co-occur within the same problem in the seed dataset. We construct a concept graph $\mathcal{G} = (\mathcal{C}, \mathcal{E})$, where nodes represent individual concepts and edge weights encode their co-occurrence strength. The edge weight between two concept nodes $u$ and $v$ is defined as:
\[
w(u, v) = \log(\mathrm{freq}(u, v) + \varepsilon),
\]
where $\mathrm{freq}(u, v)$ denotes the observed co-occurrence frequency in $\mathbf{Q}_{\text{seed}}$, and $\varepsilon$ is a small constant for numerical stability.

Given the constructed graph, we sample concept combinations using a random walk sampling. Specifically, we first uniformly sample an initial topic node, and then perform up to six steps of a random walk on $\mathcal{G}$. At each step, the transition probability from node $u$ to a neighboring node $v \in \mathcal{N}(u)$ is proportional to the corresponding edge weight:
\[
P(u,v) = \frac{\exp{w(u, v)}}{\sum_{v' \in \mathcal{N}(u)} \exp{w(u, v')}}.
\]
Each random walk episode yields a sampled concept set $s$, which is subsequently used for problem generation. Given the sampled concept set $s$, we prompt \textit{GPT-4o} to generate 20K new problems for further training.

\subsection{RL Training Configuration}

During training, we set the temperature to $0.6$, batch size to $128$, mini-batch size to $64$, and learning rate to $1e{-6}$, with a KL loss coefficient $\beta=0.005$. Unless otherwise specified, models are trained for 250 steps. For each question, we sample $8$ rollouts with a maximum response length of $16,384$ tokens. All experiments are conducted on a cluster of 8x NVIDIA A100 80G GPUs.

\section{Evaluation Details}
\label{rl-evaluation-datasets}
\subsection{Evaluation Datasets}

We use four code datasets for evaluation: LiveCodeBench, CodeContests, MBPP and CodeForces. For LiveCodeBench, we select problems from 08/01/24 to 02/01/25 to avoid overlap with the DeepCoder training dataset. For the remaining datasets, we follow \citet{DBLP:journals/corr/abs-2506-03136}'s process. Specifically, for CodeContests, we extract tasks with difficulty level $\le$ 2 and randomly generate an evaluation set of 239 examples. For MBPP, we use the standard test set. For CodeForces, we randomly sample 467 examples for evaluation.

\subsection{Sandbox Environment}
For evaluations that require code execution, we use a local sandbox implemented as a separate, guard-railed Python subprocess. Our sandbox strictly follows the official LiveCodeBench repository, ensuring consistency and fairness across evaluations.

\section{Full Results on RM-Bench}\label{appdix:rm-bench}

We evaluate \our at different scales on RM-Bench and report the full results in \autoref{tab:appendix-rmbench}. Across all model sizes, \our consistently outperforms its base \textsc{SkyworkRM} across all domains, highlighting the benefits of training on code preference data. Notably, even the smallest model, \our-1.7B, surpasses \textsc{AceCodeRM}-32B, demonstrating strong capability.

\begin{table*}[!ht]\small
\caption{Full evaluation results on RM-Bench across diverse reward models. {$\uparrow(\downarrow)$} indicates the improvement (degradation) compared to the SkyworkRM at same model size.}
\label{tab:appendix-rmbench}
\centering
\resizebox{\linewidth}{!}{%
\begin{tabular}{@{}lllllllll@{}}
\toprule
\multicolumn{1}{l}{\multirow{2}{*}{\textbf{\textsc{Models}}}} & \multicolumn{7}{c}{\textbf{\textsc{RM-Bench}}}                           & \multirow{2}{*}{\textbf{\textsc{Avg.}}} \\
\multicolumn{1}{c}{}                                        & Chat          & Math  & Code  & Safety & Hard  & Normal & Easy  &                               \\ \midrule

\textsc{AceCodeRM-7B}                                                & 66.7 & 65.3  & 66.9  & 89.9   & 62.2  & 74.4   & 79.9  & 72.2                          \\
\textsc{AceCodeRM-32B}                                               & 73.7          & 70.5  & 72.1  & 88.0     & 65.5  & 78.3   & 78.3  & 76.1                          \\  \midrule

\textsc{SkyworkRM-1.7B}                                  & {69.6}         & {71.4}    & {72.3} & {92.9}  & {54.5} & {82.3}  & {92.8} & {76.6}                         \\

\textsc{CodeScaler-1.7B}                                                & {74.4}\color{red}\textsubscript{(4.8$\uparrow$)}         & {74.7}\color{red}\textsubscript{(3.3$\uparrow$)}  & {73.1}\color{red}\textsubscript{(0.8$\uparrow$)}  & {93.1}\color{red}\textsubscript{(0.2$\uparrow$)}  & {61.5}\color{red}\textsubscript{(7.0$\uparrow$)} & {83.2}\color{red}\textsubscript{(0.9$\uparrow$)}  & {91.7}\color{ForestGreen}\textsubscript{(1.1$\downarrow$)}  & {78.8}\color{red}\textsubscript{(2.2$\uparrow$)}                          \\ \midrule
\textsc{SkyworkRM-4B} &{78.2}         & {73.6}    & {74.4} & {95.7}  & {64.4} & {85.0}  & {92.1} & {80.5}                         \\

\textsc{CodeScaler-4B}                                                & {80.4}\color{red}\textsubscript{(2.2$\uparrow$)}         & {79.0}\color{red}\textsubscript{(5.4$\uparrow$)}  & {76.3}\color{red}\textsubscript{(1.9$\uparrow$)}  & {95.8}\color{red}\textsubscript{(0.1$\uparrow$)}  & {69.2}\color{red}\textsubscript{(4.8$\uparrow$)} & {86.5}\color{red}\textsubscript{(1.5$\uparrow$)}  & {92.9}\color{red}\textsubscript{(0.8$\uparrow$)}  & {82.9}\color{red}\textsubscript{(2.4$\uparrow$)}                          \\ \midrule

\textsc{SkyworkRM-8B}                                  & {80.6}         & {75.0}    & {73.6} & {96.5}  & {67.0} & {85.5}  & {91.8} & {81.4}                         \\

\textsc{CodeScaler-8B}                                                & {83.0}\color{red}\textsubscript{(2.4$\uparrow$)}         & {79.9}\color{red}\textsubscript{(4.9$\uparrow$)}  & {76.9}\color{red}\textsubscript{(3.3$\uparrow$)}  & {96.4}\color{ForestGreen}\textsubscript{(0.1$\downarrow$)}  & {71.8}\color{red}\textsubscript{(4.8$\uparrow$)} & {87.9}\color{red}\textsubscript{(2.4$\uparrow$)}  & {92.5}\color{red}\textsubscript{(0.7$\uparrow$)}  & {84.1}\color{red}\textsubscript{(2.7$\uparrow$)}                          \\

\bottomrule
\end{tabular}
}
\end{table*}

\section{Generalization Analysis}\label{appdix:generalization}

To evaluate whether \our generalizes beyond the Qwen model family, we conduct a cross-model-family Best-of-N evaluation.

\noindent\textbf{Cross-Model-Family Transfer.} Since \our serves as a reward model whose objective is to assign fair scores to generated code, it is fundamentally model-agnostic. To empirically verify this, we evaluate several non-Qwen policy models using \texttt{BoN@8} on CodeContests. As shown in \autoref{tab:cross-family-bon}, \our consistently outperforms both \textsc{SkyworkRM} and \textsc{AceCodeRM} across both model families, confirming its cross-family scoring quality.

\begin{table}[H]
\centering
\caption{Best-of-$N$ (\texttt{BoN@8}) evaluation on CodeContests with non-Qwen policy models. {$\uparrow(\downarrow)$} indicates the improvement (degradation) compared to the Pass@1.}
\label{tab:cross-family-bon}
\begin{tabular}{llcl}
\toprule
\textbf{Policy} & \textbf{Reward Model} & \textbf{Pass@1} & \textbf{BoN@8} \\
\midrule
\multirow{3}{*}{DeepSeek-Coder-6.7B-Instruct} & \textsc{SkyworkRM}   & \multirow{3}{*}{18.41} & 21.76 \\
 & \textsc{AceCodeRM}      &  & 23.85 \\
 & \our &  & \textbf{25.10}\color{red}\textsubscript{(6.69$\uparrow$)} \\
\midrule
\multirow{3}{*}{Llama-3.1-8B-Instruct}        & \textsc{SkyworkRM}   & \multirow{3}{*}{15.90} & 19.67 \\
 & \textsc{AceCodeRM}      &  & 20.08 \\
 & \our &  & \textbf{21.76}\color{red}\textsubscript{(5.86$\uparrow$)} \\
\bottomrule
\end{tabular}
\end{table}

\section{Training Stability Analysis}\label{appdix:training-stability}

We provide three diagnostic analyses to demonstrate training stability of \our.

\begin{figure}[H]
\centering
\includegraphics[width=\linewidth]{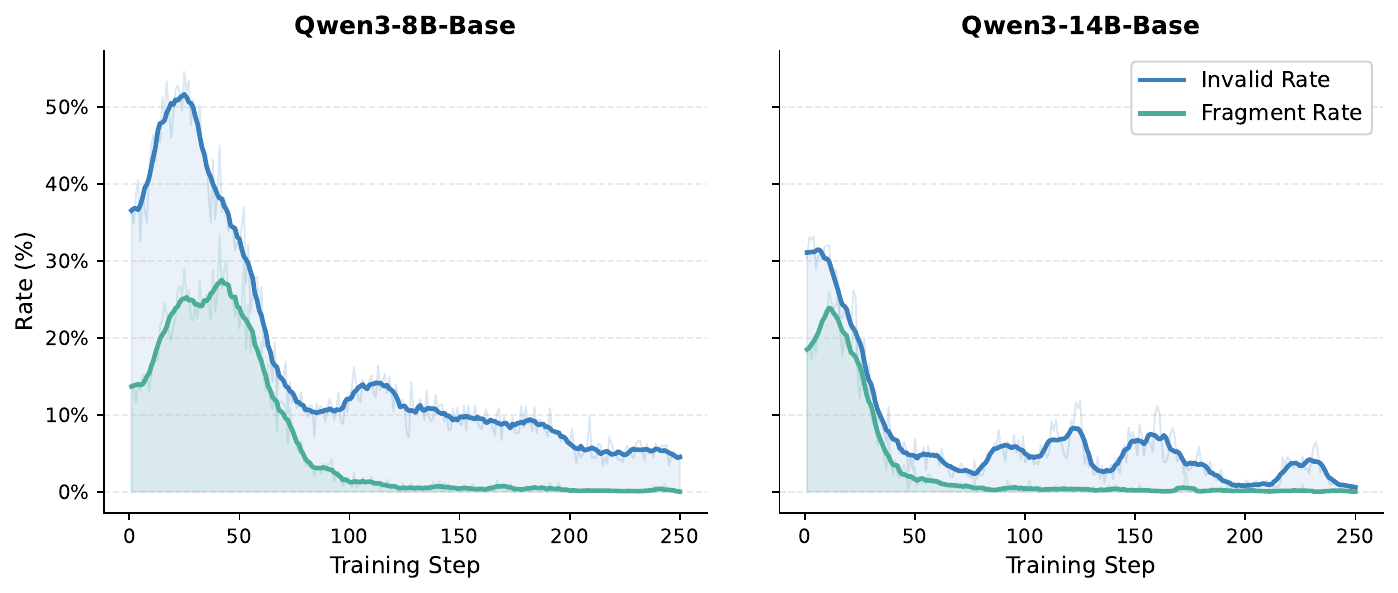}
\caption{Training dynamics of invalid rate and fragment rate for \textsc{Qwen3-8B-Base} and \textsc{Qwen3-14B-Base} trained with \our on DeepCoder. Both rates decrease steadily after initial exploration, demonstrating effective suppression of failure modes.}
\label{fig:training-dynamics}
\end{figure}

\noindent\textbf{Training Dynamics.} \autoref{fig:training-dynamics} shows the invalid rate and fragment rate for both \textsc{Qwen3-8B-Base} and \textsc{Qwen3-14B-Base} trained with \our on DeepCoder. Both rates initially increase during early exploration but then decrease steadily. The fragment rate converges to near zero ($\leq$0.5\% by step 150), and the invalid rate drops from 30--37\% to $\sim$1.3\% (14B) and $\sim$7\% (8B) by step 250, demonstrating that the syntax-aware code extraction and validity-preserving reward shaping effectively suppresses both failure modes throughout training.

\begin{figure}[H]
\centering
\includegraphics[width=\linewidth]{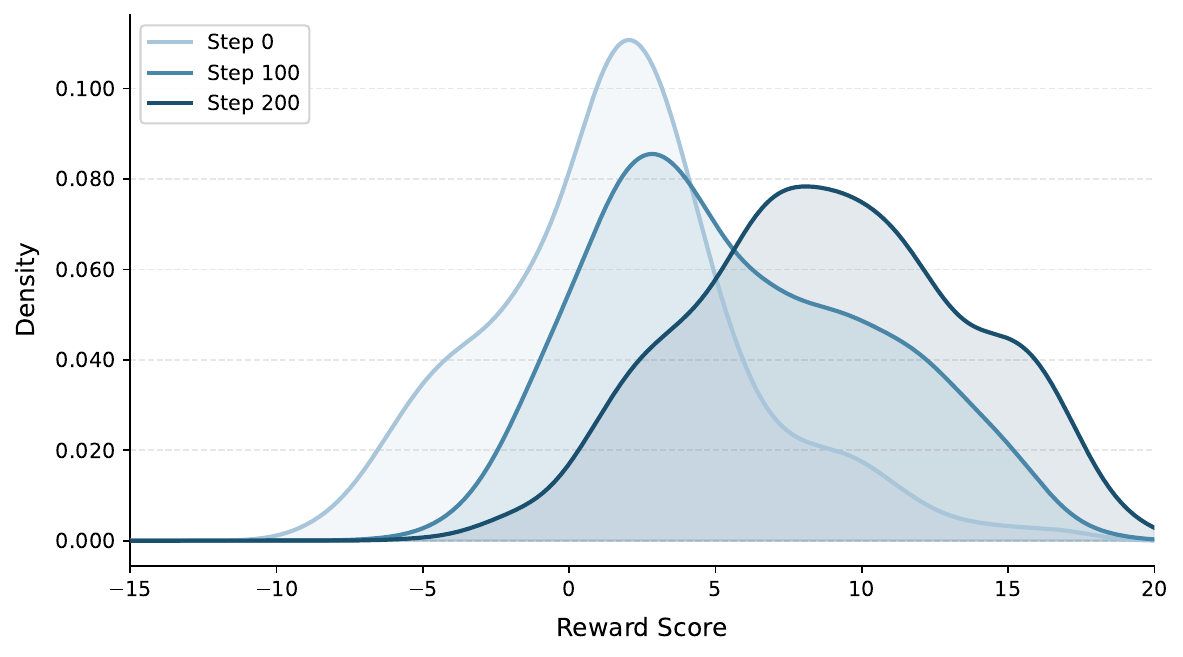}
\caption{Evolution of the RM score distribution at training steps 0, 100 and 200 (\textsc{Qwen3-8B-Base} on DeepCoder). The distribution shifts rightward steadily with no signs of reward hacking.}
\label{fig:reward-dist}
\end{figure}

\noindent\textbf{Reward Distribution Evolution.} \autoref{fig:reward-dist} illustrates the reward score distribution at training steps 0, 50, 100, 150, and 200 for \textsc{Qwen3-8B-Base} on DeepCoder. The distribution shifts steadily rightward as training progresses: the mean score rises from 1.73 to 8.82, and the proportion of negative-reward rollouts drops from 32.2\% to 2.1\%.

\begin{figure}[H]
\centering
\includegraphics[width=\linewidth]{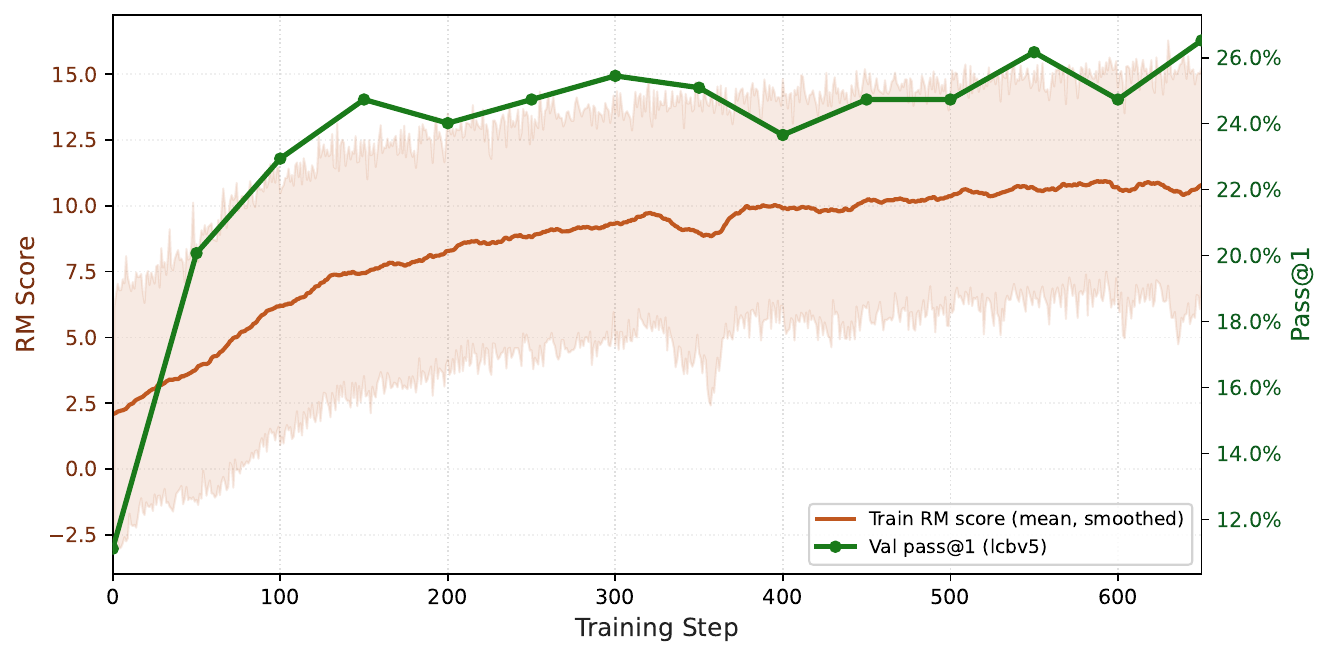}
\caption{Extended training of \textsc{Qwen3-8B-Base} on DeepCoder from 250 to 650 steps. Both RM score and pass@1 increase steadily with no divergence, confirming the absence of reward hacking.}
\label{fig:reward-hacking}
\end{figure}

\noindent\textbf{Extended Reward Hacking Analysis.} To verify that reward hacking does not occur under prolonged training, we extend the \textsc{Qwen3-8B-Base} experiment on DeepCoder from 250 to 650 steps, evaluating every 50 steps. As shown in \autoref{fig:reward-hacking}, both RM score and pass@1 increase steadily throughout training, providing empirical evidence against reward hacking over extended training horizons.

\section{Sensitivity to Misalignment Ratio}\label{appdix:misalignment-sensitivity}

To improve the robustness of \our, we augment the preference pairs by adding misaligned pairs where the code solution is generated for a different problem (see \autoref{appendix:rm-training-details}). We investigate the sensitivity of \our to the misalignment augmentation ratio by training three variants with 0\%, 30\%, and 50\% augmentation, and evaluating each on a held-out test set containing both aligned and misaligned samples.

\begin{table}[H]
\centering
\caption{Sensitivity analysis of the misalignment augmentation ratio. The 0\%$\to$30\% jump yields a large improvement in discrimination (AUROC 0.911$\to$0.990), while the 30\%$\to$50\% gain is marginal, confirming that 30\% is a robust and sufficient augmentation ratio.}
\label{tab:misalign-ratio}
\begin{tabular}{lccc}
\toprule
\textbf{Misalignment Ratio} & \textbf{Score on Misaligned Code} & \textbf{Margin (pos$-$neg)} & \textbf{AUROC} \\
\midrule
0\%  & $-6.87  \pm 2.71$ & $15.42 \pm 5.69$ & 0.911 \\
\textbf{30\%} & $-14.85 \pm 2.31$ & $23.40 \pm 5.37$ & \textbf{0.990} \\
50\% & $-17.45 \pm 2.20$ & $25.68 \pm 6.11$ & 0.990 \\
\bottomrule
\end{tabular}
\end{table}

Without augmentation, \our assigns only mildly negative scores ($-6.87$) to misaligned code, with an AUROC of 0.911. With 30\% augmentation, the score drops to $-14.85$ and AUROC rises sharply to 0.990, indicating that the model has internalized a strong notion of problem-code alignment. Increasing the ratio to 50\% yields only marginal further improvement, confirming that the alignment signal saturates around 30\%.

\section{Failure Case Analysis}\label{appdix:failure-case-analysis}

To understand \our's failure modes, we conduct a quantitative and qualitative analysis using training trajectory records from \textsc{Qwen3-8B-Base} on DeepCoder with ground-truth execution labels. We identify two failure types: \textit{False Positives} (FP), where the highest-scored rollout in a group fails execution, and \textit{False Negatives} (FN), where the lowest-scored rollout passes execution. Groups with uniform outcomes are excluded, yielding 347 FP and 166 FN cases.

Comparing surface features across groups reveals two systematic biases:

\noindent\textbf{Length Bias (FP vs.\ TP).} Despite being incorrect, FP code is 28\% longer than TP code, contains 37\% more algorithmic keywords, and is accompanied by 38\% longer chain-of-thought responses. This suggests that \our partially relies on response length and keyword density as quality proxies.

\begin{figure}[H]
\centering
\includegraphics[width=\linewidth]{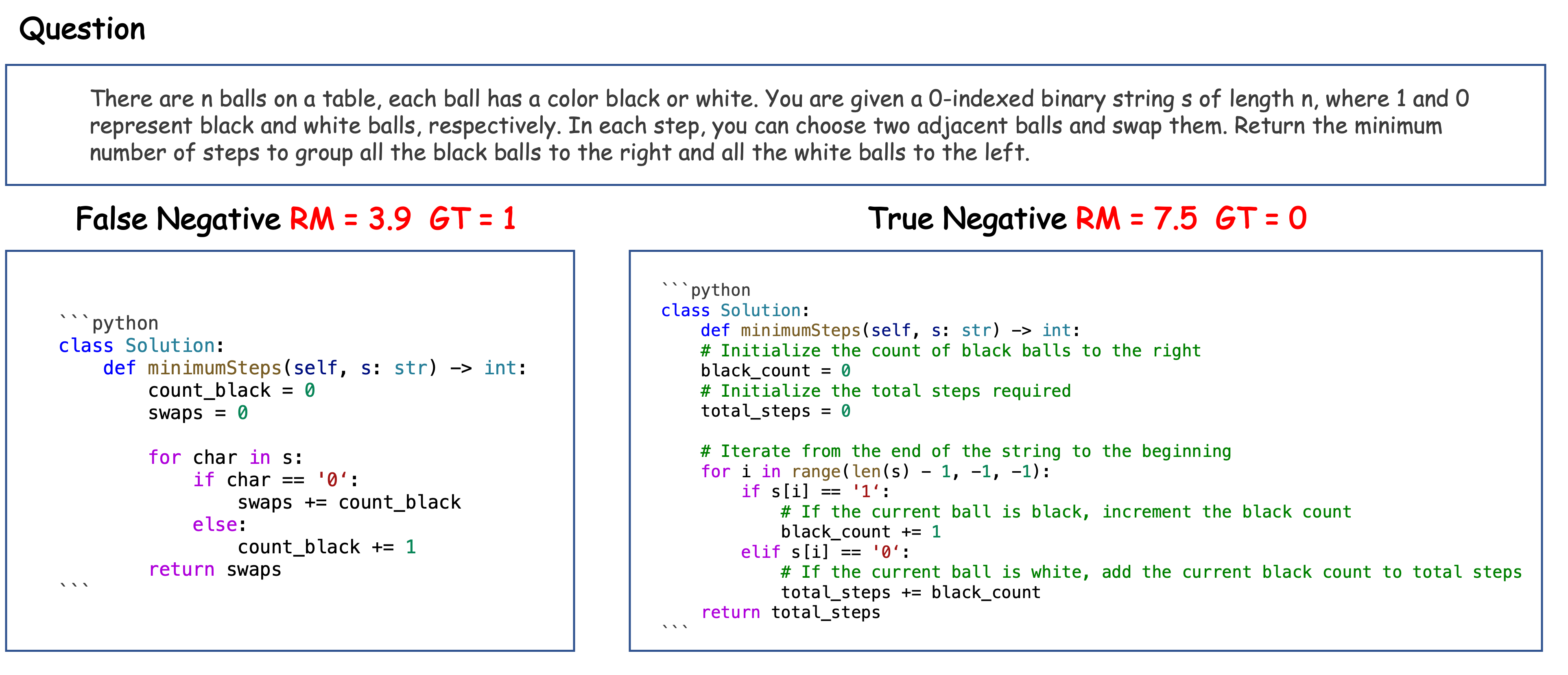}
\caption{Both solutions implement an O(n) single-pass approach. The FN (correct, 278 chars) sweeps left-to-right, counting for each white ball the number of black balls to its left. The TN solution (incorrect, 661 chars) mirrors the traversal right-to-left, inadvertently solving the opposite goal (blacks left, whites right). Despite being 2.4× shorter and correct, the FN receives RM = 3.969 vs. RM = 7.531 for the wrong solution.}
\label{fig:failure-case-study}
\end{figure}

\noindent\textbf{Brevity Under-Valuation (FN vs.\ TN).} Correct but low-scored FN responses are $2\times$ shorter than incorrect TN responses. FN code also shows higher rates of modular multi-function design (+32\%) and recursion (+18\%), suggesting that concise, well-structured solutions are systematically undervalued. \autoref{fig:failure-case-study} illustrates a representative case.

\section{Prompt Template}

\begin{tcolorbox}[breakable,colback=red!5!white,colframe=red!75!black,title=Answering Prompt Template]
You are an expert Python programmer. You will be given a question (problem specification) and will generate a correct Python program that matches the specification and passes all tests.
\newline\newline
\{\{question\}\}
\newline\newline
\#\#\# Format: Read the inputs from stdin solve the problem and write the answer to stdout (do not directly test on the sample inputs). Enclose your code within delimiters as follows. Ensure that when the python program runs, it reads the inputs, runs the algorithm and writes output to STDOUT.
\newline\newline
\texttt{```}python

\# YOUR CODE HERE

\texttt{```}
\newline\newline
\#\#\# Answer: (use the provided format with backticks)

\end{tcolorbox}

\begin{tcolorbox}[breakable,colback=red!5!white,colframe=red!75!black,title=Concept Extraction Template]
Act as a Programming Teacher and analyze the provided question. Start by identifying 1 or 2 general topics that a student is being assessed on. Structure your response as:

``Topics:

1. \texttt{<}Topic 1\texttt{>}

2. \texttt{<}Topic 2\texttt{>}"
\newline\newline
Next, highlight 1 to 5 specific knowledge points that the question evaluates. Structure your response as:

``Specific Knowledge Points:

1. \texttt{<}Knowledge Point 1\texttt{>}

2. \texttt{<}Knowledge Point 2\texttt{>}

3. \texttt{<}Knowledge Point 3\texttt{>}

4. \texttt{<}Knowledge Point 4\texttt{>}

5. \texttt{<}Knowledge Point 5\texttt{>}"
\newline\newline
The topics and specific knowledge points should be terms that are concise and commonly used in academia or industry.
\newline\newline
\#\#\# Provided question:

\{\{question\}\}
\newline\newline
\#\#\# Analysis:

\end{tcolorbox}

\begin{tcolorbox}[breakable,colback=red!5!white,colframe=red!75!black,title=Problem Generation Prompt Template]

Act as a Programming Teacher and create a new question and its solution based on the provided topics and knowledge points. Ensure that the created questions:

1. Adhere to the provided topics.

2. Necessitate the combined use of the associated knowledge points.
\newline\newline
\# Example

Topics:

1. Graph Theory

2. Combinatorics

Knowledge Points:

1. Graph Coloring

2. Connected Components

3. Dynamic Programming

4. Modular Arithmetic

Created Question:

You are given a grid, consisting of $2$ rows and $n$ columns. Each cell of this grid should be colored either black or white.

Two cells are considered neighbours if they have a common border and share the same color. Two cells $A$ and $B$ belong to the same component if they are neighbours, or if there is a neighbour of $A$ that belongs to the same component with $B$.

Let's call some bicoloring beautiful if it has exactly $k$ components.

Count the number of beautiful bicolorings. The number can be big enough, so print the answer modulo $998244353$.
\newline\newline
-----Input----

The only line contains two integers $n$ and $k$ ($1 \le n \le 1000$, $1 \le k \le 2n$) the number of columns in a grid and the number of components required.
\newline\newline
-----Output----

Print a single integer — the number of beautiful bicolorings modulo $998244353$.
\newline\newline
-----Examples----

Input

34

Output

12

Input

41

Output

2

Input

12

Output

2
\newline\newline
Topics:

1. String Manipulation

Knowledge Points:

1. Understanding and manipulating string data structures

2. Dynamic Programming
\newline\newline
Try to create a question for the last one. Structure your response as:

Created Question:

\texttt{<}Question\texttt{>}

\end{tcolorbox}

\section{A Case Study on Flawed Synthesis Test Cases}\label{case-study}

Taking a synthesis problem from rStarCoder as an example, this task requires computing the number of valid arrangements that contain exactly $n$ successive flowers of the same type. Although the problem description specifies an input range of $3\leq n\leq30$, the synthesized test cases only cover values up to $n=9$. This limitation likely arises because the mutual verification mechanism fails to reach consensus on certain corner or large-input cases, which are consequently discarded during test case construction. As a result, the synthesized test suite is incomplete, and the missing cases are often the most challenging ones that truly distinguish efficient solutions from inefficient ones.

In this setting, it is possible to construct a flawed solution that relies on a naive pure-recursive enumeration strategy without any form of memoization or dynamic programming. The recursive function \texttt{count\_ways\_pure\_recursive} explores all possible color assignments at each position, leading to an exponential-time complexity with respect to the flowerbed length $2n-2$. While this approach is sufficient for small values of $n$ (e.g., $n\leq 9$), it becomes prohibitively expensive as $n$ increases.

Consequently, this solution passes all synthesized test cases and is incorrectly regarded as correct. However, when evaluated on larger inputs within the valid constraint range, it exhibits severe performance degradation and ultimately results in time limit exceeded (TLE) errors.

\begin{tcolorbox}[breakable,colback=blue!5!white,colframe=blue!75!black,title=rStarCoder synthetic\_rl\_testcase synthetic\_50341]\label{flaw-synthesis}

Problem Description:

A city council wants to arrange a flowerbed in a park to contain exactly n successive flowers of the same type. There are four types of flowers available for planting.

The flowerbed consists of one line of (2n - 2) spaces. The total number of flowers is greater than the number of spaces available. Additionally, the number of flowers of each type exceeds the number of spaces on the flowerbed. Determine the number of ways to arrange the flowerbed so that it contains exactly n successive flowers of the same type.

Input Format:

An integer n (3 $\leq$ n $\leq$ 30) representing the number of successive flowers of the same type.

Output Format:

An integer representing the number of ways to arrange the flowerbed in the required manner.

Examples

Input:

3

Output:

24

Input:

4

Output:

132

\tcblower
Inputs:

[3,4,5,6,7,8,9]

Outputs:

[24,132,672,3264,15360,70656,319488]
\end{tcolorbox}

\newtcolorbox{codebox}{
  colback=gray!5,
  colframe=gray!40,
  boxrule=0.5pt,
  arc=2pt,
  left=3pt,
  right=3pt,
  top=4pt,
  bottom=4pt,
  breakable,
  title=A flawed solution to synthetic\_50341 yet passes all test cases
}

\begin{codebox}\label{flaw-solution}
\begin{lstlisting}[language=Python,basicstyle=\ttfamily\footnotesize,breaklines=true,breakatwhitespace=true]
import sys

def solve():
    line = sys.stdin.readline()
    if not line:
        return
    n = int(line.strip())

    total_len = 2 * n - 2

    def count_ways_pure_recursive(idx, current_streak, found_exactly_n):
        if current_streak > n:
            return 0

        if idx == total_len:
            return 1 if (found_exactly_n or current_streak == n) else 0

        res = 0
        res += count_ways_pure_recursive(idx + 1, current_streak + 1, found_exactly_n)

        if current_streak == n:
            res += 3 * count_ways_pure_recursive(idx + 1, 1, True)
        else:
            res += 3 * count_ways_pure_recursive(idx + 1, 1, found_exactly_n)

        return res

    final_ans = 4 * count_ways_pure_recursive(1, 1, False)
    print(final_ans)

if __name__ == "__main__":
    solve()
\end{lstlisting}
\end{codebox}

\end{document}